\RequirePackage{snapshot}
\documentclass[10pt,letterpaper]{article}
\usepackage[letterpaper, total={6in, 9in}]{geometry}
\usepackage{times}
\usepackage{epsfig}
\usepackage{graphicx}
\usepackage{amsfonts}
\usepackage{amsmath}
\usepackage{amssymb}
\usepackage{adjustbox}
\usepackage{booktabs}
\usepackage{multirow}
\usepackage{array}
\usepackage{comment}
\usepackage{color}
\usepackage{algorithm2e}
\usepackage[utf8]{inputenc}
\usepackage[T1]{fontenc}
\usepackage{cancel}
\usepackage{bm}
\usepackage{caption}
\usepackage{cuted}
\usepackage{microtype}
\usepackage{nicefrac}
\usepackage{pbox}
\usepackage{setspace}
\usepackage{siunitx}
\usepackage{subfigure}
\usepackage{url}
\usepackage{varwidth}
\usepackage{xspace}
\usepackage{xcolor}
\usepackage{mathtools}

\DeclareMathOperator*{\argmin}{arg\,min}
\DeclarePairedDelimiter{\ceil}{\lceil}{\rceil}

\newcommand{\R}{\mathbb{R}}

\usepackage[pagebackref=true,breaklinks=true,letterpaper=true,colorlinks,bookmarks=false]{hyperref}
\usepackage{cleveref}
\newcommand\Ccancel[2][black]{\renewcommand\CancelColor{\color{#1}}\xcancel{#2}}

\usepackage{capt-of,etoolbox}
\makeatletter
\apptocmd\@maketitle{{\myfigure{}\par}}{}{}
\makeatother

\begin{document}

\title{RidgeSfM: Structure from Motion via Robust Pairwise Matching Under Depth Uncertainty}

\newcommand\myfigure{%
\vspace{-5mm}
\centering\includegraphics[width=1\textwidth]{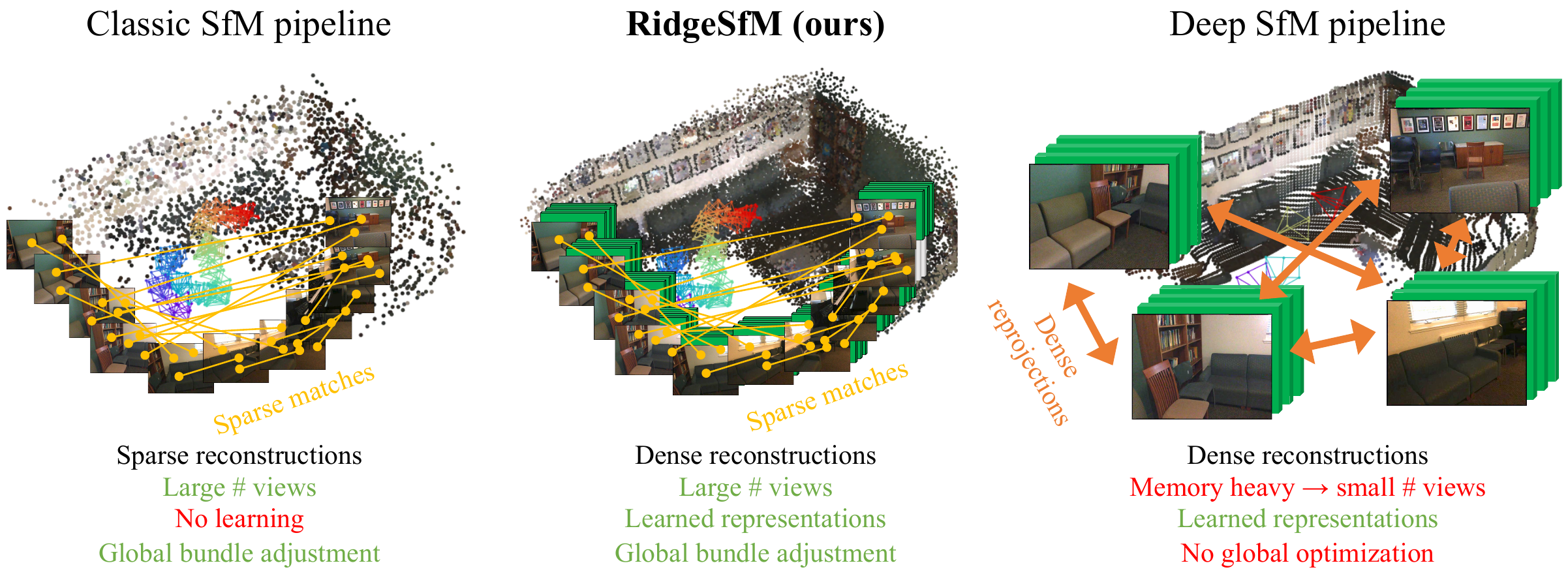}
\captionof{figure}{\textbf{RidgeSfM efficiently marries the classic SfM pipelines with recent deep approaches.} Like classic SfM, it employs a large scale bundle adjustment over hundreds of frames. Similar to deep methods, it is capable of harnessing geometry priors from a large dataset with geometric annotations.
We release the source code at \url{https://github.com/facebookresearch/RidgeSfM}.
}
\label{f:splash}
\vspace{10mm}
}

\author{%
  Benjamin Graham \hspace{0.5cm} David Novotny\\
  Facebook AI Research \\
  London \\
  \texttt{\{benjamingraham,dnovotny\}@fb.com} \\
}
\date{}
\maketitle

\begin{abstract}
  We consider the problem of simultaneously estimating a dense depth map and camera pose for a large set of images of an indoor scene. While classical SfM pipelines rely on a two-step approach where cameras are first estimated using a bundle adjustment in order to ground the ensuing multi-view stereo stage, both our poses and dense reconstructions are a direct output of an altered bundle adjuster. To this end, we parametrize each depth map with a linear combination of a limited number of basis "depth-planes" predicted in a monocular fashion by a deep net. Using a set of high-quality sparse keypoint matches, we optimize over the per-frame linear combinations of depth planes and camera poses to form a geometrically consistent cloud of keypoints. 
  Although our bundle adjustment only considers sparse keypoints, the inferred linear coefficients of the basis planes immediately give us dense depth maps. RidgeSfM is able to collectively align hundreds of frames, which is its main advantage over recent memory-heavy deep alternatives that can align at most 10 frames.
  Quantitative comparisons reveal performance superior to a state-of-the-art large-scale SfM pipeline. 
\end{abstract}

\section{Introduction}

Estimating the 3D structure and camera motion from image sequences is a traditional task that attracted the computer vision community since its inception.
Modern Structure-from-Motion (SfM) systems \cite{schoenberger2016sfm,schoenberger2016mvs} are robust and able to reconstruct thousands if not millions of photos from significantly heterogeneous image collections.
Suprisingly, since the the seminal works from \cite{frahm2010building,agarwal2011building,pollefeys2008detailed,snavely2006photo}, there has been little change to the fundamentals of the SfM pipeline.

This comes as an even bigger surprise after deep learning revolutionized most of the classic CV tasks - SfM did not enjoy the benefits of deep learning to the extent other subfields have, and classic SfM building blocks have prevailed. While there have been many efforts to boost reconstruction algorithms with deep learning \cite{demon,tateno2017cnn,bloesch2018codeslam,zhou2017unsupervised,DeepV2D}, due to their memory requirements, they only consider small-scale setups with a handful of images that are incomparable to the vast scenes that classic SfM bundle adjusters can process. Omitting the global optimization step constitutes a significant drawback since any short-term tracking system will eventually drift without loop closure.

In this paper, we aim at achieving a more harmonious marriage between deep learning (DL) and the classic SfM pipelines. Departing from the standard DL approach which considers losses defined over dense pixel-wise predictions, we tap into the classic idea of utilizing only sparse keypoint matches, as their low memory footprint is the main enabler of global optimization. However, we still employ CNNs in order to learn powerful priors from annotated data.

The crux of our method lies in predicting the \emph{allowed factors of variation} of 3D positions of image points. More specifically, instead of employing the standard direct monocular regression of depth for each image, we task our deep network to predict an intermediate representation of dense depth in the form of a set of basis "depth-planes" that span the modes of ambiguity of the true image depth. Importantly, our per-frame depth maps are simple linear combinations of the basis planes, bringing several benefits that summarize our contributions:

First, the optimized depth prediction is constrained to lie on a compact manifold represented with a small number of scalar coefficients of the basis planes. This alleviates the need for ad-hoc depth regularizers, such as TV-norms.

Second, the linearity of our representation allows us to optimize reprojection losses for only a small set of sparse keypoints in each frame without the need to keep the entire basis depth planes and intermediate CNN features in memory. This brings tremendous memory savings and allows us to run bundle adjustment at a similar scale to classic SfM pipelines. 

\begin{figure*}[t] \centering
\includegraphics[width=1\textwidth]{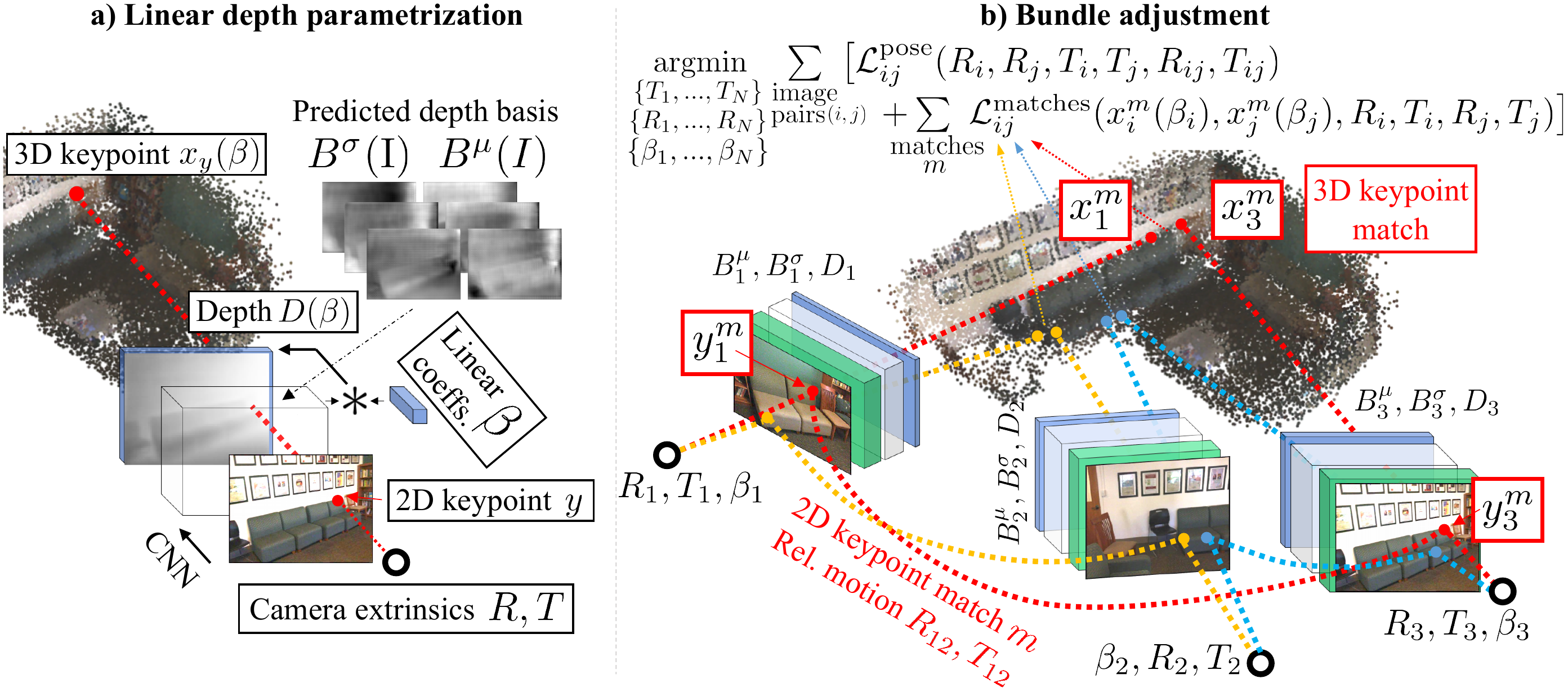}
\caption{
\textbf{An overview of RidgeSfM.}
(a) The key component is an efficient linear parametrization of depth maps
using a linear combination of basis depth planes $B^\mu(I), B^\sigma(I)$ (predicted by a CNN) with coefficients $\beta$.
(b) The linear parametrization allows to execute a memory efficient bundle adjustment (BA)  that only considers sparse keypoint matches and, thus, can align thousands of frames. 
Once BA optimizes the coefficients $\beta$ and extrinsics $R, T$, we can efficiently recover dense depth $D = B^\mu(I)+B^\sigma(I)\beta$.
Our BA thus indirectly optimizes over dense depth.
}
\end{figure*}

Third, once our BA finishes, we can apply the returned basis coefficients to the depth planes to obtain the dense depth maps in a straightforward fashion. This is more efficient than classic SfM pipelines which require additional multi-view stereo processing to recover dense reconstructions.

Our empirical evaluation on the ScanNet dataset reveals that RidgeSfM outperforms a popular representative of a classic SfM pipeline (COLMAP \cite{schoenberger2016sfm}) in a large-scale global adjustment regime.
Its performance is on par with deep memory-heavy alternatives for pair-wise image matching.

\section{Related Work}

\paragraph{Structure-from-Motion}
Structure-from-Motion (SfM) constitutes the most classic line of work that targets recovering the 3D structure of a scene and tracking of the camera. 
Starting from the early works that focused on limited numbers of images \cite{mohr1995relative,beardsley19963d}, modern SfM pipelines evolved into mature systems capable of reconstructing thousands \cite{heinly2015reconstructing,radenovic2016dusk,schonberger2015single,wu2013towards,frahm2010building,agarwal2011building}
if not millions of photos of various in/outdoor scenes. 
A particularly popular SfM pipeline that effectively combines the fundamental findings from the body of previous work, COLMAP, was built by Schoenberger et al. \cite{schoenberger2016sfm,schoenberger2016mvs}. 
It follows the nowadays standard design pattern: 
1) Geometrically verified keypoint matches are established between pairs of images. 
2) The estimated matches and relative camera motions are fed into an incremental ``bundle adjustment'' (BA) that globally optimizes the camera positions and triangulates a sparse 3D point cloud of the scene.
3) Multi-View Stereo utilizes the inferred absolute cameras to produce dense depth maps.

\paragraph{SLAM}
Related to SfM are SLAM methods that aim at real-time tracking of a moving camera. PTAM \cite{klein2007parallel} was one of the first practical systems that allowed real-time tracking and mapping using a pair of reconstruction and tracking threads. 
PTAM was later extended to dense reconstructions in DTAM \cite{newcombe2011dtam}. 
LSD-SLAM \cite{engel2014lsd} is another notable example of a method capable of semi-dense reconstruction and tracking. 
Finally, DSO \cite{engel2017direct} attained a good trade-off between speed and accuracy by directly optimizing photometric error evaluated at sparse keypoints.

The aforementioned classic SLAM and SfM systems are carefully ``hand-engineered'' methods that, despite being the current methods of choice in practice, have a limited ability to leverage priors learnable from large geometry-annotated datasets. 
The next paragraph discusses methods that constitute promising future learning-based directions.

\paragraph{Deep learning of geometry.}
The success of deep learning brought an expected invasion of deep networks to the SfM/SLAM domain.
Initial approaches have focused solely on monocular depth estimation \cite{eigen2015predicting,laina2016deeper,wang2015towards,li2015depth,liu2015deep}, or on estimating the camera pose \cite{agrawal2015learning,kendall2015posenet,jayaraman2015learning,kanezaki2018rotationnet,wang2017deepvo}. Deep CNNs were also leveraged to describe image pixels for better matching in standard SfM pipelines \cite{zagoruyko2015learning,dosovitskiy2015discriminative,zbontar2015computing,taira2018inloc}. 

However, the most relevant approaches focus on reconstruction of both ego-motion and depth.
DeMoN \cite{demon} predicts disparities with the FlowNet architecture \cite{ilg2017flownet} to  ground its predictions. BA-Net, DeepTAM and LS-Net \cite{clark2018learning,banet,zhou2018deeptam} proposed iterative architectures capable of geometrically aligning a pair of images. 
CNN-SLAM \cite{tateno2017cnn} and DVSO \cite{yang2018deep} studied the use of deep monocular depth predictors for improving the performance of existing SLAM pipelines.
More recently, 3DVO \cite{yang2020d3vo} proposed an architecture that allowed to handle several types of reconstruction ambiguities. 
Several methods have also explored unsupervised learning of depth and ego-motion from videos \cite{zhou2017unsupervised,yin2018geonet,godard2019digging,godard2017unsupervised,wang2018learning}.

Notably, CodeSLAM \cite{bloesch2018codeslam} and BA-Net \cite{banet} are similar to our method in the parametrization of depth maps using a latent code which is later refined with the cameras.

Unfortunately, all the aforementioned deep methods suffer from large memory consumption which prevents executing a global bundle adjustment over thousands of frames. This is because they consider dense reconstruction errors that, apart from limiting applicability to small camera motions, require complicated decoding or matching networks to be stored in GPU memory at reconstruction time. In constrast, RidgeSfM optimizes all scene cameras jointly because it restricts its optimization to a set of sparse matches in the bundle. Importantly, although we consider losses evaluated at sparse landmarks, the linearity of our latent depth parametrization allows us to simultaneously solve for the dense depth of each scene image.

\paragraph{Image keypoints.} Sparse set of keypoints that can be matched across images are a crucial building block of the standard SfM pipeline. Classic examples include SIFT \cite{sift} and ORB \cite{orb}. Deep learning approaches include LF-Net \cite{LFNet}, D2-Net \cite{D2Net}, R2D2 \cite{r2d2}, and SuperPoint \cite{superpoint}.

\section{Task and naming conventions}

Given a set of images of a scene 
$\{I_i | I_i \in \R^{3 \times H \times W}\}_{i=1}^N$ of height $H$ and width $W$, 
the goal of our work is estimating the parametrization of the absolute orientation (extrinsics) of the per-frame cameras
$\{(R_i, T_i) | R_i \in SO(3), T_i \in \R^3\}_{i=1}^N$ 
as well as the depth maps 
$\{(D_i | D_i \in \R^{H \times W}\}_{i=1}^N$.

We will follow the ensuing convention. 
By sampling $D_i$ at a pixel location 
$y \in \{ 1, ..., W \} \times \{1,...,H\}$
we can identify $y$'s depth value $d_y \in \R$. 
Per-pixel depth, together with the calibration matrix 
$K_i \in \R^{3\times3}$ of $I_i$'s camera, allows to back-project each pixel $y$ to its 
corresponding 3D point $x(d_y) = K_i^{-1} d_y [y_{[1]},y_{[2]},1]^T$
in the camera coordinates.
Here $z_{[k]}$ is an operator that retrieves $k$-th value of a vector $z$.
The camera calibration matrices $K_i$ are assumed to be known.
A point $x^c \in \R^3$ in the coordinates of camera $i$ is mapped to scene coordinates $x^w \in \R^3$ with $x^w = R_i x^c + T_i$.

\section{Depth parametrization} \label{s:depth_param}

For SfM pipelines, the parametrization of the extrinsics $\{R_i, T_i\}$ is straightforward, but the same cannot be claimed for the per-frame depths $D_i$. Since the set of plausible depth maps forms a low dimensional manifold, 
measures have to be taken to ensure compactness of the representation of $D_i$. 

A classic solution is to employ a regularizer, such as TV-norm, that ensures spatial smoothness. Such regularizer typically entails a cumbersome hyperparameter tuning and only indirectly enforces the depth maps to follow their natural manifold.
Furthermore, these depth regularizers are hand-engineered functions that do not allow to learn priors from datasets with geometric annotations.

In order to deal with the latter, recently, Bloesch et. al \cite{bloesch2018codeslam} proposed to parametrize depths with a trained deep non-linear mapping $\phi_{\text{CodeSLAM}}(\beta_i) = D_i$, where $\beta_i \in \R^{K}$ is a low dimensional latent depth code. While this greatly improves the generated depth maps, a major disadvantage is the substantial memory footprint due to $\phi_{\text{CodeSLAM}}$ being a heavy de-convolutional network, which has to be evaluated during every step of the SfM optimization.

\paragraph{Parametrizing depth with mean and factors of variation.}
In order to deal with the aforementioned issues and obtain a learnable memory-efficient latent depth parametrization, RidgeSfM, similar to \cite{banet}, parametrizes depth maps in a linear fashion as a weighted combination of depth basis planes. Our parametrization function $\phi$ takes the form:\footnote{
Note that we reshape depth $D_i\in\R^{H\times W}$ to a vector $\text{vec}(D_i)\in\R^{HW}$ to simplify the notation.
}
\begin{equation}\label{e:depth_linear_param}
\text{vec}(D_i) = \phi(\beta_i, I_i) = B^\mu(I_i) + B^\sigma(I_i) \beta_i,
\end{equation}
where $\beta_i \in \R^{K}$ is a set of linear coefficients that are used to adjust $D_i$. The predicted mean depth $B^\mu(I_i) \in \R^{H W}$ and factors of variation $B^\sigma(I_i) \in \R^{K \times H W}$ are the outputs of a small convolutional network $B$.

This formulation has two main advantages: 
1) After $B^\mu(I_i)$ and $B^\sigma(I_i)$ are predicted from $I_i$, one can erase the intermediate tensors of network $B$ from memory and alter the depth $D_i$ by only optimizing over $\beta_i$.
2) As $K \ll HW$, our depth code $\beta_i$ forms a compact geometric bottleneck that is unlikely to result in unnatural depth values: we can optimize the coefficients for a sparse set of keypoints and then safely extrapolate to optimized dense depth maps.

\paragraph{Learning depth uncertainty with Ridge Regression Loss.}
Our deep network $B$ is related to modern deep monocular depth predictors. 
While the inherent ambiguity of monocular depth estimation is considered as a limiting factor that makes the problem underconstrained and thus hard to solve, the linearity of \cref{e:depth_linear_param} allows us to easily train the network $B$. 

Given a training image $I$ annotated with a ground truth depth map $D^\star$, we exploit the monocular depth prediction ambiguity by letting our network predict the basis vectors $B^\mu(I)$, $B^\sigma(I)$ and solving for a $\beta^\star$ that leads to a good approximation of $D^\star$ by employing Ridge Regression:
\begin{align}\label{e:ridge_regression}
\beta^\star 
&= \argmin_{\beta \in \R^K}  \|\phi(\beta,I)-D^\star\|_2^2 +\lambda \|\beta\|_2^2  \\
&= (B^\sigma(I)^\intercal B^\sigma(I) + \lambda \mathbf{I}_K)^{-1} 
B^\sigma(I)^\intercal (D^\star - B^\mu(I)).\nonumber
\end{align}
Given $\beta^\star$, we train our network $B$ to minimize the following loss function:
\begin{align}\label{e:ridge_regression_loss}
    \mathcal{L}_\text{depth}&(I | B, D^\star) =  \\
        & \|B^\mu(I)-D^\star\|_2^2 
        + \|B^\mu(I) + B^\sigma(I) \beta^\star - D^\star \| \nonumber \\
        & + \lambda\|\beta^\star\|_2^2
        + \|\text{RowVar}(B^\sigma(I)) - 1\|_1,\nonumber
\end{align}
where $\text{RowVar}(B^\sigma(I))$ denotes the sample variance of the rows of $B^\sigma(I)$. 
This encourages the network to put the best-guess prediction $B^\mu(I)$ close to the ground truth, to construct $B^\sigma(I)$ such that $D^\star \approx D=\phi(\beta^\star, I)$ with $\|\beta^\star\|_2^2$ small.
This way, the basis $B^{\mu/\sigma}(I)$ is predicted such that it spans the most significant modes of uncertainty of the depth map $D$ for image $I$.

\paragraph{Training data and network architecture.}
The depth prediction network $B$ is trained using batched gradient descent to optimize the loss 
$\mathcal{L}_\text{depth}$ averaged over a large dataset of images and depth maps. Here we use 1412 scenes from the ScanNet RGBD dataset for training. The architecture of $B$ is based on U-Net \cite{ronneberger2015u}, built with inverted residual building blocks \cite{InvertedResidual}. 
Full details are in the supplementary.

\section{Ridge Structure from Motion}

The input to RidgeSfM is a sequence of images. For each image $I_i$, we extract a sparse set of keypoint location $y_j\in I_i$, and a corresponding collection of feature vectors. We also use the trained depth network to predict the depth and factors of variation $B^{\mu/\sigma}(I_{i})$.

RidgeSfM, similar to classic SfM pipelines, then reconstructs scenes in two steps.
First, egomotions between pairs of frames are estimated and then a global bundle adjustment procedure is carried out.

\subsection{Pairwise RidgeSfM}\label{s:rsfm_pairwise}
Given a pair of images $(I_i, I_j)$, we use the values of $B^{\mu/\sigma}(I_{i/j})$ at the keypoint locations to estimate the corresponding relative camera motion $R_{ij} \in SO(3), T_{ij} \in \R^3$.

\paragraph{Weakly verified matches.}
We create set of weakly verified matches 
$\mathcal{M}_{ij} = \{(y_i^m, y_j^m) | y_i^m \in I_i, y_j^m \in I_j\}_{m=1}^M$
as follows. We first search for nearest-neighbor pairs of keypoints in feature space. 
We then remove any pairs that fail the crosscheck critera, or that are considered outliers by OpenCV's \verb findFundamentalMat  LMedS function.

\paragraph{Pairwise RidgeSfM alignment of matched keypoints.}
Let $b_i^{\mu m}(I_i) \in \R$ and $b_i^{\sigma m}(I_i) \in R^K$ denote per-pixel basis vectors obtained by sampling $B^\mu(I_i)$ and $B^\sigma(I_i)$ respectively at location $y_i^m$.
Given $b_i^{(\mu/\sigma) m}$, we can obtain the per-pixel depth with $d_i^m(\beta_i, I_i) = \beta_i^\intercal b_i^{\sigma m}(I_i) + b_i^{\mu m}(I_i)$. 
Finally, $d_i^m(\beta_i, I_i)$ is used to backproject pixel $y_i^m$ to its 3D camera coordinates with 
$x_i^m(\beta_i, I_i) = K_i^{-1} d_i^m(\beta_i, I_i) \big[y_i^m, 1]^T$.
Then, pairwise RidgeSfM solves for $\{R_{ij}, T_{ij}, \beta_i, \beta_j\}$
by minimizing the pairwise alignment loss $\mathcal{L}_{\text{pw}}$:
\begin{align}
\mathcal{L}_{\text{pw}}(\mathcal{M}_{ij}) = 
\sum_{m=1}^M  \ell_{ij}^m 
+ \lambda( \| \beta_i \|^2_2 + \| \beta_j \|^2_2); \nonumber  \\
\ell_{ij}^m = \| R_{ij} x_i^m(\beta_i, I_i) + T_{ij} - x_j^m(\beta_j, I_j) \|^2_2, \label{e:pw_ridgesfm}
\end{align}
We minimize $\mathcal{L}_{\text{pw}}$ with coordinate descent by alternating two steps until convergence:
1) Given $\beta_i, \beta_j$ we solve for $R_{ij}, T_{ij}$ using Umeyama's rigid alignment algorithm \cite{umeyama};
2) Given $R_{ij}, T_{ij}$ \cref{e:pw_ridgesfm} reduces to a simple ridge regression problem which allows to solve for $\beta_i, \beta_j$. 
The algorithm is initialized with $\beta_i=\beta_j=\mathbf{0}$, corresponding to the depth network's predicted mean depth.

\paragraph{Egomotion estimation by progressive growing of matches.}
The 3D alignment procedure is not robust to outliers, so we use RANSAC to build a subset $\mathcal{M}^I_{ij}\subset \mathcal{M}_{ij}$ of strongly geometrically verified inlier matches.

Starting with $\mathcal{M}^I_{ij}$ consisting of $M=3$ matches randomly drawn from $\mathcal{M}_{ij}$, we alternate between 
(i) optimizing $\mathcal{L}_{\text{pw}}(\mathcal{M}_{ij}^I)$, and 
(ii) increasing the number of active matches $M$ by a multiplicative factor: we increase $M$ by re-selecting $\mathcal{M}_{ij}^I$ to be the subset of matches $\mathcal{M}_{ij}$ that are most closely aligned (i.e. their $\ell^m_{ij}$ is small) given the current estimate of $\{R_{ij},T_{ij},\beta_i,\beta_j\}$. 
The process stops when the alignment error $\max_{m \in \mathcal{M}_{ij}^I} \ell^m_{ij}$ exceeds a precision threshold. Full details are in supplementary material.

\subsection{Bundle-adjustment with RidgeSfM} \label{s:rsfm_bundle}

Given a procedure for pairwise matching that returns sets of inlier matches $\mathcal{M}^I_{ij}$ as well as relative camera motions, we propose to use a bundle adjuster to collectively estimate absolute camera orientations and dense depths for all images in a scene, as detailed below.

Broadly, we minimize the following bundle adjustment loss $\mathcal{L}^{\text{bundle}}$:
\begin{align}\label{e:bundle_adjust}
\min_{\{R_i\},\{T_i\}, \{\beta_i\}}
\sum_{(i,j) \in \mathcal{I}} &
\mathcal{L}^\text{matches}_{ij}(R_i, R_j, T_i, T_j, \beta_i, \beta_j) \nonumber \\
&+ \mathcal{L}^\text{pose}_{ij}(R_i, R_j, T_i, T_j),
\end{align}  
where the outer sum is carried over the set $\mathcal{I}$ 
of all pairs of images with a significant set of verified matches $\mathcal{M}^I_{ij}$.
We represent the absolute rotations $R_i$ as cumulative products of rotations each stored in Tait-Bryan angles; and translations $T_i$ as cumulative sums of 3-dimensional vectors.

The first term $\mathcal{L}^\text{matches}_{ij}$, which enforces consistency of the 3D locations of matched points, is defined as follows:

\begin{align}\label{e:bundle_adjust_match}
    \mathcal{L}^\text{matches}_{ij} &= 
    \sum_{m \in \mathcal{M}^I_{ij}}
    \sigma(u_{ij}^m) ~ e_{ij} + \lambda_u ~ \sigma(-u_{ij}^m); \\
    e_{ij} &= \|
        [R_i  x_i^m(\beta_i, I_i) + T_i ]
        - 
        [R_j x_j^m(\beta_j, I_j) + T_j ]
    \|_2,\nonumber 
\end{align}
where $\sigma$ is the logistic function, $u_{ij}^m \in \R$ are auxiliary variables that limit the effect of bad matches, and $\lambda_u=0.3$ is the strength of the regularizer $\sigma(-u_{ij}^m)$ preventing the trivial solution of $\sigma(u_{ij}^m) = 0$ everywhere.
Importantly, since the number of matches between pairs of images is significantly lower than the number of image pixels, $\mathcal{L}^\text{matches}_{ij}$ 
can be integrated over a large number of image pairs $\mathcal{I}$
allowing to optimize $\beta_i$, and conversely dense depths $D_i$,
in a large-scale regime.

The second term $\mathcal{L}^\text{pose}_{ij}$ aids the convergence of the scene cameras.
More specifically, it minimizes the discrepancy between the absolute camera orientations $(R_i,T_i)$ and and the relative cameras poses $(R_{ij},T_{ij})$ predicted by pairwise RidgeSfM:
\begin{equation}\label{e:bundle_adjust_pose}
\mathcal{L}^\text{pose}_{ij} =
\|R_i - R_j R_{ij} \|_1 + 
\|T_i - R_j T_{ij} - T_j \|_1.
\end{equation}

\paragraph{Optimization.}
The Adam optimizer is used, with weight decay applied to the $\beta_i$.
To improve convergence, we first minimize the $\mathcal{L}^\text{pose}_{ij}$ losses incrementally for $6|\mathcal{I}|$ iterations;
at step $t$ we optimize their partial sum over the first $\min(\lceil t/5 \rceil,|\mathcal{I}|)$ elements of $\mathcal{I}$, sorted by $j$.
We then optimize the full scene loss \cref{e:bundle_adjust} until convergence.

\begin{table*}[t]
    \centering{}
    \begin{tabular}{l|cccc|cccc}
    \toprule
    Method 
    & \multicolumn{4}{c|}{COLMAP SfM pipeline}
    &\multicolumn{4}{c}{RidgeSfM using SuperPoint features}
    \\
    \midrule
    Skip rate 
    & 1 & 3 & 10 & 30 
    & 1 & 3 & 10 & 30\\
    \midrule   
Camera rotation (degrees)
&22.12 &11.17 &9.85 &29.85
&7.09 &7.84 &7.35 &12.84
\\
Camera center (m)
&0.973 &0.597 &0.540 &1.085
&0.296 &0.314 &0.331 &0.489
\\
Depth map $L_1$ err. (m)
&0.941 &0.763 &0.727 &1.184
&0.221 &0.234 &0.243 &0.322
\\
Depth map RMSE (m)
&1.138 &1.012 &0.990 &1.386
&0.305 &0.332 &0.343 &0.432
\\
PCL $L_1$ err. (m)
&0.647 &0.642 &0.639 &0.860
&0.209 &0.258 &0.303 &0.454
\\
PCL RMSE (m)
&0.821 &0.885 &0.906 &1.081
&0.289 &0.345 &0.393 &0.569
\\
\midrule
  Successful reconstructions 
  & 99\%        &  100\%       &   98\%        &    81\% 
  & 100\%       &  100\%       &  100\%        &   100\%   
  \\
    \bottomrule
    \end{tabular}
    \caption{
    \textbf{Quantitative comparison with COLMAP on large-scale bundle adjustment on the ScanNet dataset}. For COLMAP, evaluation is based on the available reconstructed frames for scenes where reconstruction was at least partially successful. For RidgeSfM, the evaluation uses all frames in and all scenes.
    \label{tbl:colmap}
    }
    \end{table*}
  
\begin{table*}[t]
    \centering{}%
    \begin{tabular}{l|cccc|cccc}
    \toprule
    Ablation 
& \multicolumn{4}{c|}{$\mathcal{L}^{\text{bundle}}=\sum \Ccancel[red]{\mathcal{L}^\text{matches}_{ij}}+ \mathcal{L}^\text{pose}_{ij}$}
& \multicolumn{4}{c}{\Ccancel[red]{SuperPoint} SIFT features} \\
\midrule
    Skip rate 
    & 1 & 3 & 10 & 30
    & 1 & 3 & 10 & 30\\
    \midrule   
Camera rotation (deg.)
&7.22 &9.14 &9.10 &25.05 
& 7.78&  9.1 &  9.08& 19.16\\
Camera center (m)
&0.306 &0.344 &0.395 &0.853
&0.318   &0.377    &0.420    &0.676
\\
Depth map $L_1$ err. (m)
&0.236 &0.250 &0.269 &0.382
&0.229   &0.259    &0.274    &0.422
\\
Depth map RMSE (m)
&0.331 &0.357 &0.379 &0.493
&0.314   &0.359    &0.376    &0.533
\\
PCL $L_1$ err. (m)
&0.224 &0.297 &0.386 &0.881
&0.224   &0.309    &0.387    &0.677
\\
PCL RMSE (m)
&0.307 &0.383 &0.474 &1.017
&0.304   &0.398    &0.478    &0.811
\\
\bottomrule
\end{tabular}
\caption{
  \textbf{Ablation study on ScanNet. Left:} Result using only the pairwise pose loss $\mathcal{L}^\text{pose}$, rather than the full bundle adjustment loss  $\mathcal{L}^{\text{bundle}}$. 
  \ \textbf{Right:} Results for RidgeSfM but with SuperPoint features replaced with SIFT features.\label{tbl:ablation}
  }
 \end{table*}

\section{Experiments} \label{s:exp}

In this section we quantitatively and qualitatively evaluate our method.
Starting with a description of the utilized benchmark, we then present experiments
evaluating the global bundle adjustment procedure which collectively aligns hundreds of frames from an indoor scene. 
Since virtually all existing deep alternatives, such as \cite{banet,zhou2018deeptam,bloesch2018codeslam,yang2020d3vo}, do not allow for such large-scale evaluation due to their ample memory-consumption, for completeness, we compare to these method on a small-scale task of aligning image pairs.

\paragraph{Benchmark dataset.}
ScanNet \cite{scannet} is a dataset of RGB videos frames with matching depth maps, camera locations, and camera intrinsics, captured with a hand-held scanning device.
There are 1513 training scenes, and 100 test scenes.
We use the first 1412 scenes from the training set to train the depth prediction network.
We evaluate RidgeSfM on the remaining scenes that are not seen during training.
We consider the supplied camera poses as `ground truth', as they were calculated using 
the sensor depth-supervised bundle adjustment.

\paragraph{Evaluation of bundle adjustment.}
To test RidgeSfM on large sequences of images, we selected sequences of up to 300 images from the ScanNet test videos.
For each of the 100 test scenes, we picked random starting points, and sampled every $k$-th frame with the skip rate $k=1,3,10$ and 30, for a total of 400 test cases.
RidgeSfM is compared to COLMAP \cite{schoenberger2016sfm}, a popular SfM pipeline that is widely considered as the current state-of-the-art.

Using RidgeSfM we reconstruct the camera poses and dense depth maps as explained in \Cref{s:rsfm_bundle}. We use SuperPoint \cite{superpoint} as the keypoint detector. 
For COLMAP, we first run the sparse reconstruction that tracks the cameras,  and then dense depths are estimated with COLMAP's multi-view stereo method. Since COLMAP sometimes fails to estimate camera pose or depth for an image, we exclude these cases from the evaluation. Note that RidgeSfM does not enjoy this benefit of being able to exclude ambiguous frames from its evaluation; by design it is forced to reconstruct all pixels, in all frames, for every scene.

For evaluation purposes, given depth maps and camera extrinsics, we generate a dense point cloud of each scene, which is later aligned with the ground truth using Umeyama's algorithm \cite{umeyama} that estimates a 7 d.o.f. similarity transformation. After alignment and rescaling, we report several errors: Camera rotation / center error denotes the average rotation / distance between the ground truth camera's rotation / translation matrix and the prediction. The depth map $L_1$/RMSE denote errors between the ground truth and the estimated depth. Point-cloud (PCL) $L_1$/RMSE are similar to the latter and compute the distance between the per-pixel 3D coordinates obtained by backprojecting estimated depth using the estimated camera location.

Table \ref{tbl:colmap} demonstrates that RidgeSfM reconstructions are superior to COLMAP in all metrics and test scenes. \Cref{f:bundle_qual}, and videos in the supplementary material, qualitatively evaluates our reconstructions.

\begin{figure*}
\centering
\includegraphics[width=0.84\textwidth]{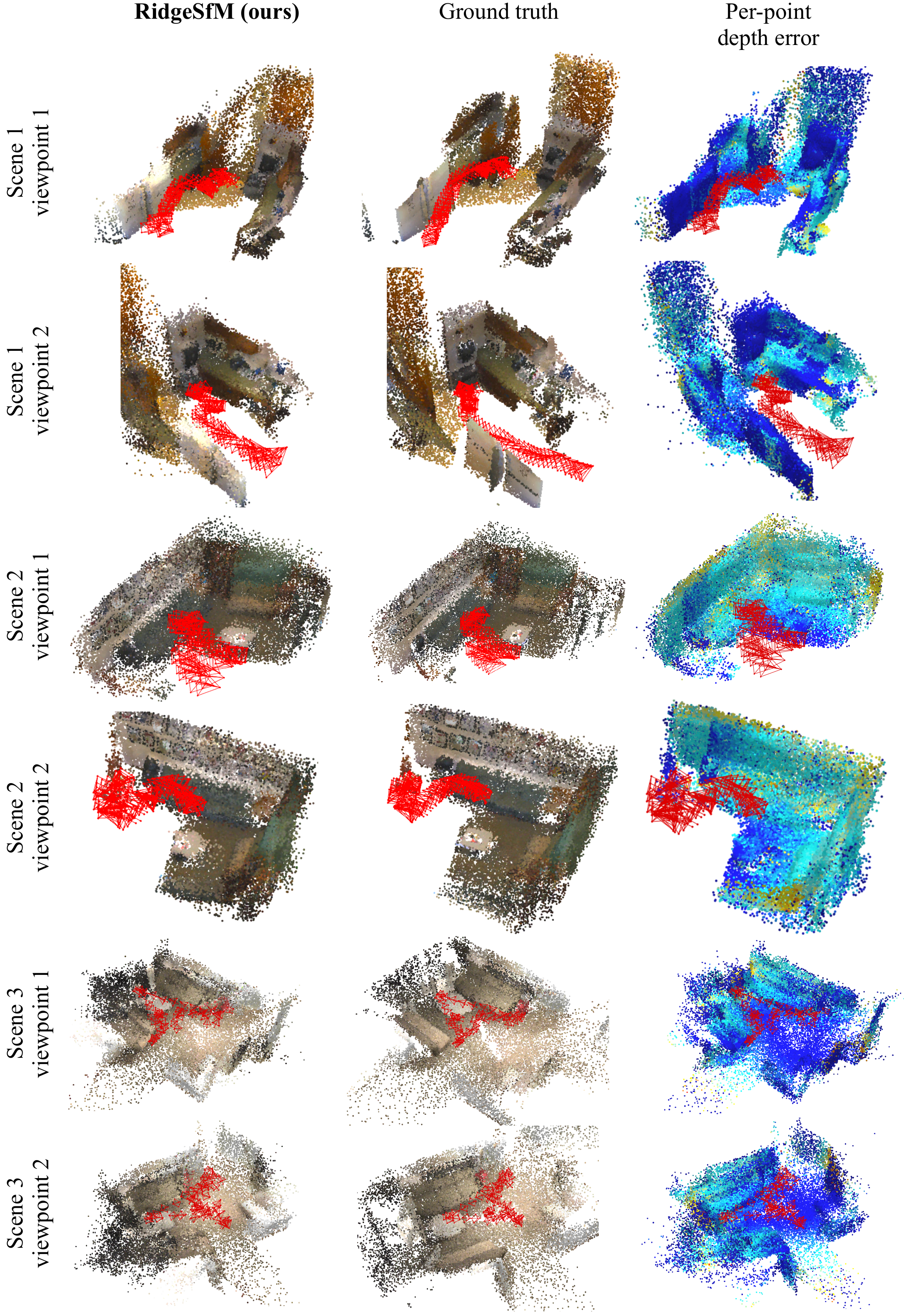}
\caption{Qualitative results comparing the dense scene-wise reconstruction and camera poses of RidgeSfM (1st column) and the ground truth (2nd column).
The 3rd column colors each point of the RidgeSfM reconstruction proportionally to its depth error (red-highest, blue-lowest). Each of the 3 depicted scenes is visualised from 2 different viewpoints.
\label{f:bundle_qual}}
\end{figure*}  

\begin{figure*}[ht]
    \newcommand{\figh}{3.5cm}
    \newcommand{\figw}{0.32\linewidth}
    \newcommand{\figcol}[2]{
        \begin{minipage}[c]{\figw}%
        \centering%
        \includegraphics[height=\figh]{pairs_superpoint/#1}\\
        \includegraphics[height=\figh]{pairs_superpoint/#2}%
        \end{minipage}%
    }
    \centering%
    \figcol{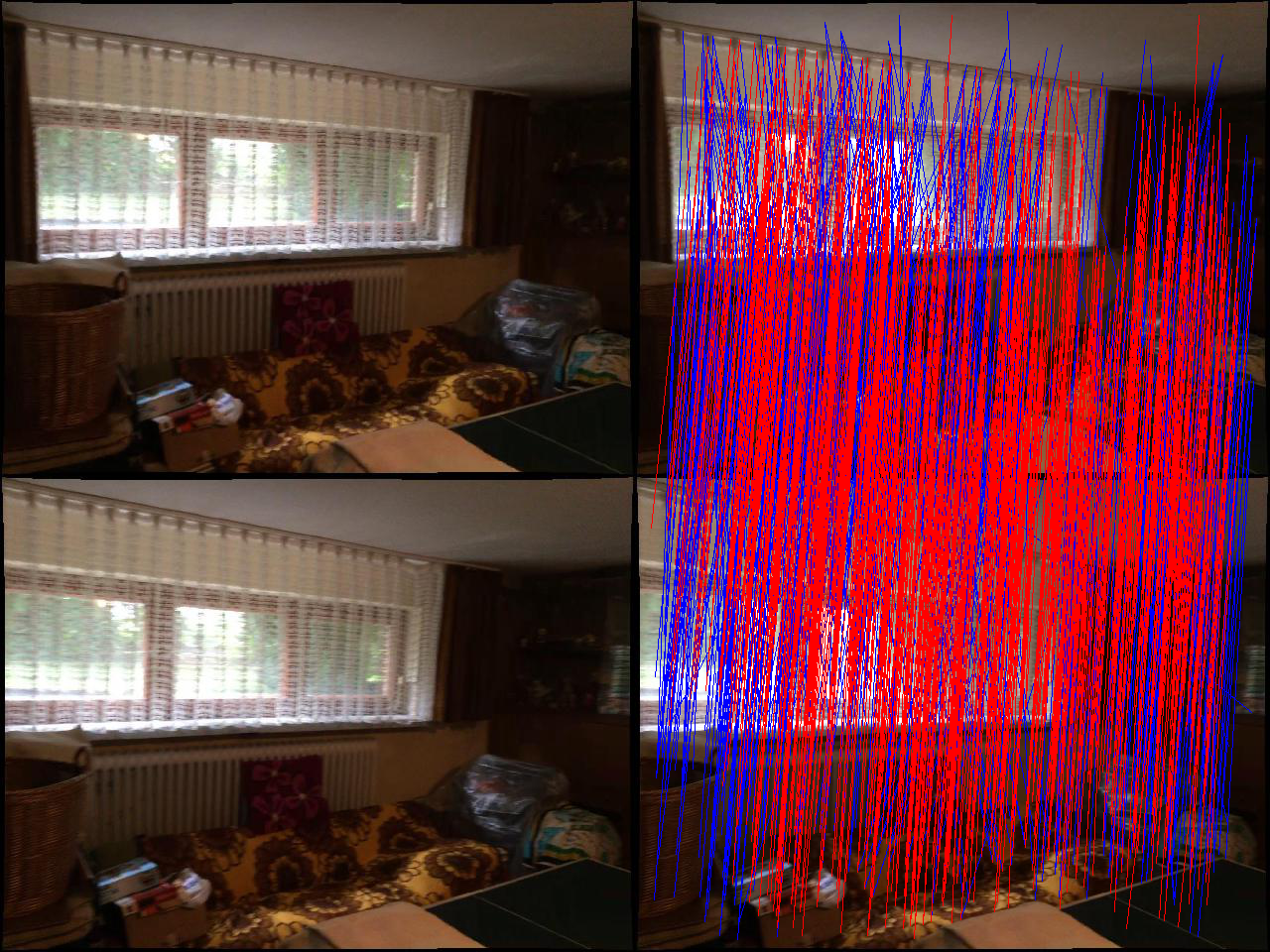}{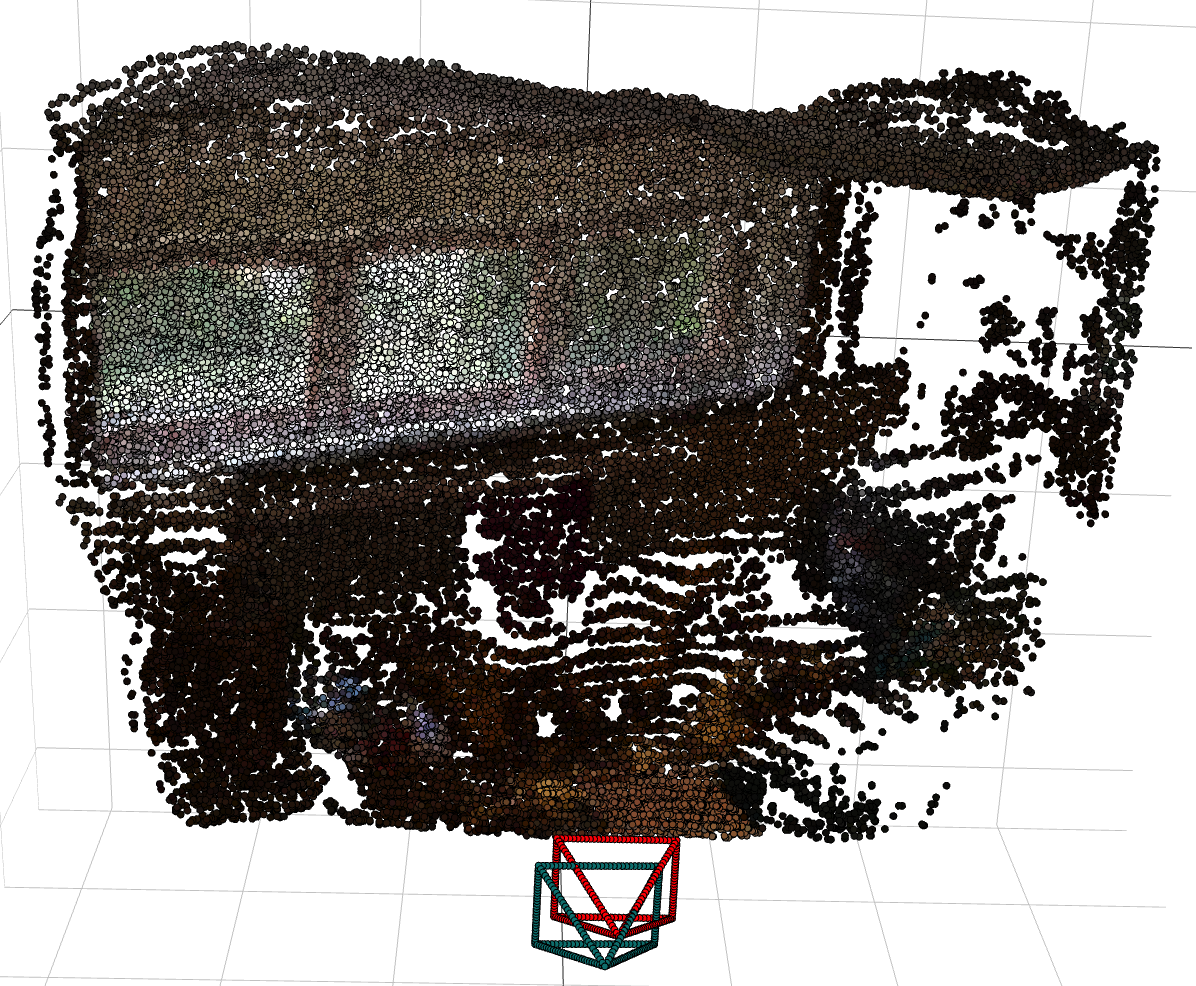}%
    \figcol{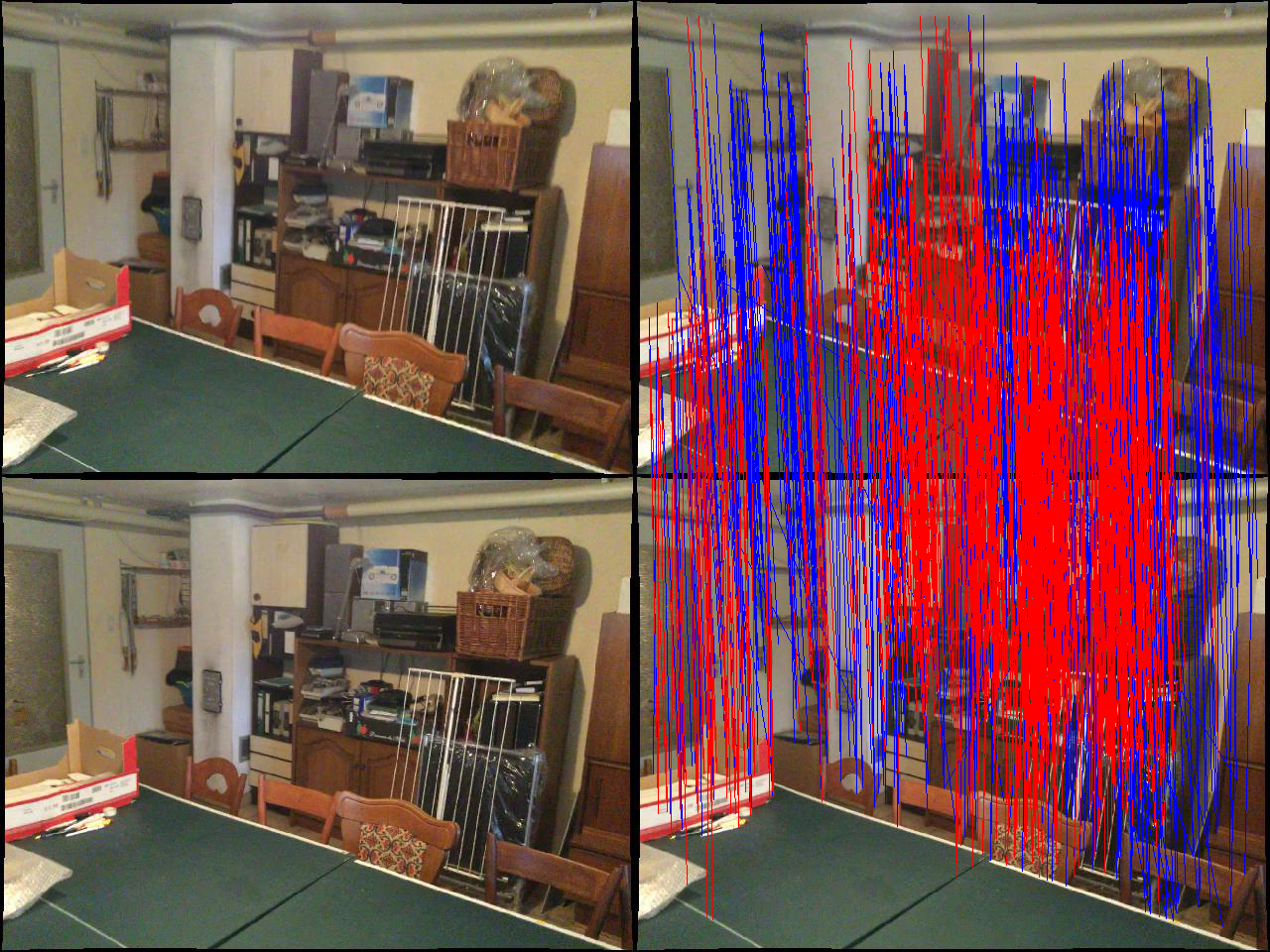}{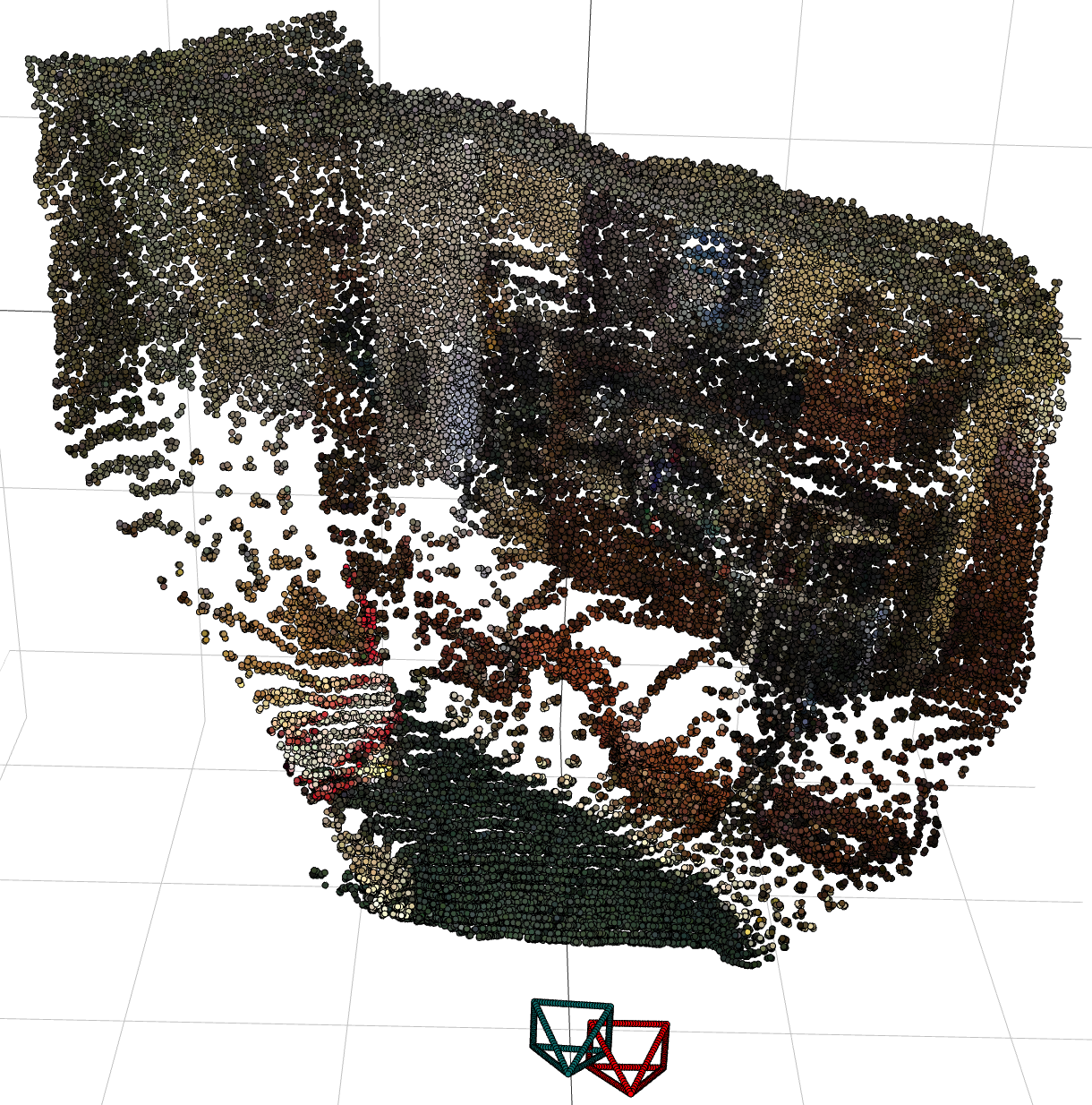}%
    \figcol{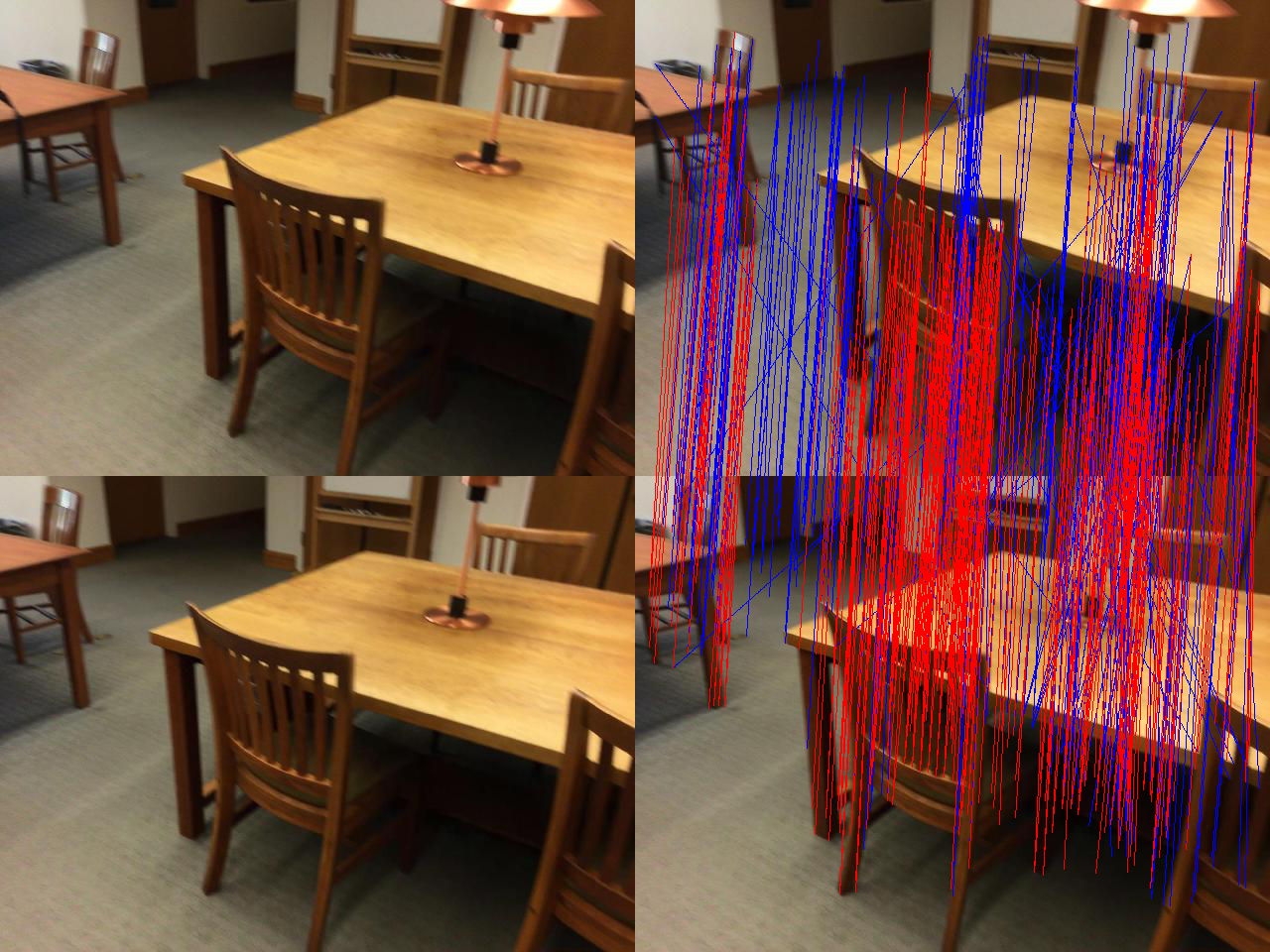}{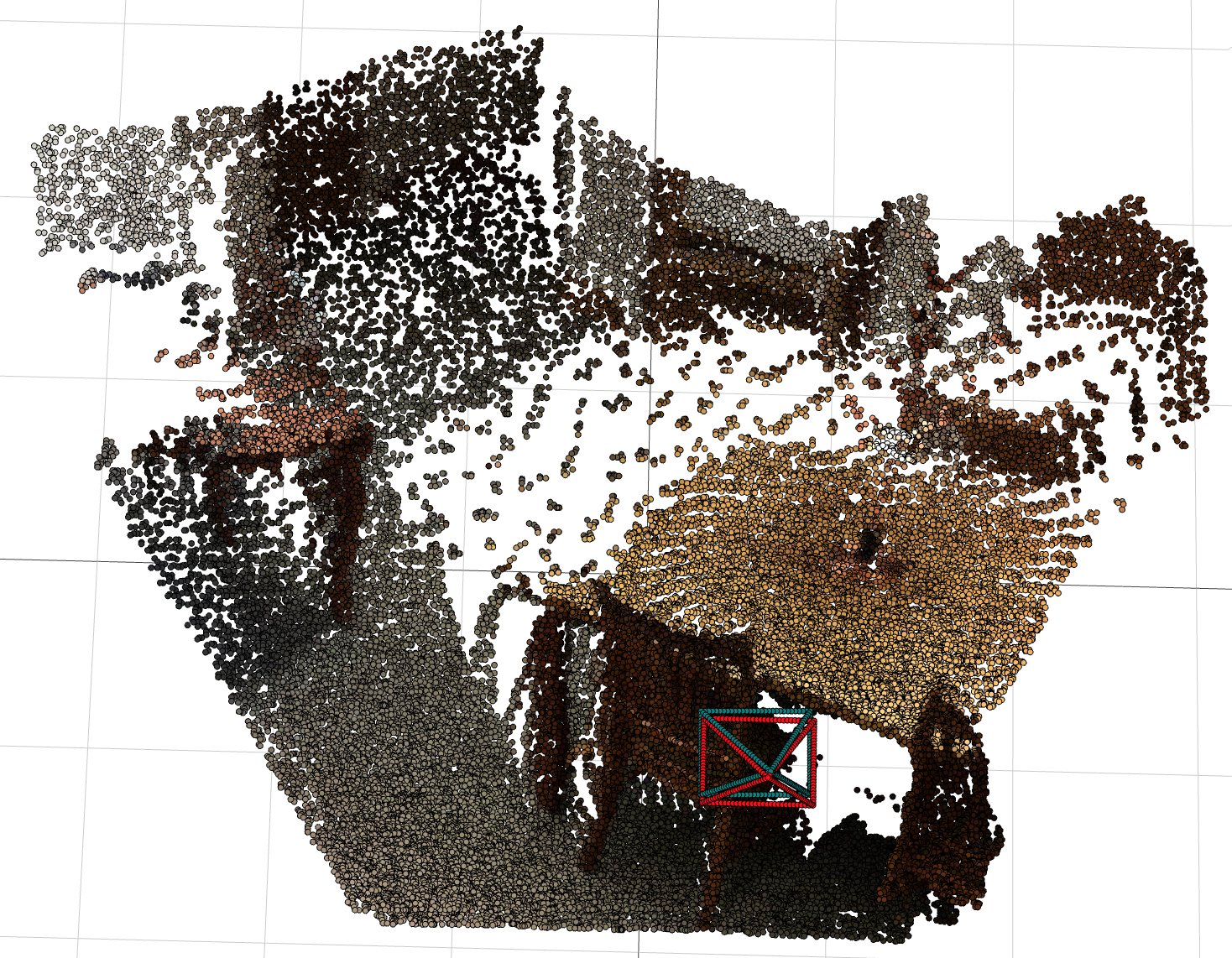}%
      \caption{
        \textbf{Example results on pairwise matching.}
        Top: Pairs of images with per-pixel correspondences. 
        Bottom: The inferred scene point clouds and cameras - we have plotted 10\% of the pixels.
        The blue lines show initial feature matches. Red matches denote the inliers of the Pairwise RidgeSfM alignment.
        \label{fig:pairs}
    }
\end{figure*}

\begin{table*}
\begin{adjustbox}{center}
    \begin{tabular}{lcccccc}
    \toprule
    Error                 & RidgeSfM   & BA-Net$^{*}$ \cite{banet} & DeMoN$^{*}$ \cite{demon} & Photometric BA$^{*}$ & Geometric BA$^{*}$ & DeepV2D $\dagger$  \cite{DeepV2D}\\
    \midrule
    Depth RMSE (m)        &  0.33        & 0.35                      & 0.76                     & 0.79                & 0.88              & 0.17 \\
    \midrule
    Rotation (degree)     &  1.94        & 1.02                      & 3.79                     & 4.41                & 8.56              & 0.63 \\
    Translation (cm)      &  7.70        & 3.39                      & 15.50                    & 21.40               & 37.00             & 1.37 \\
    \bottomrule
    \end{tabular}
\end{adjustbox}
\caption{Pairwise alignment results on the ScanNet image pairs dataset \cite{banet}. Columns with $*$ are from \cite{banet}; columns with $\dagger$ use the ground-truth median depth to scale each of the predicted depth maps.
    \label{tbl:pairs} 
    }
\end{table*}

\paragraph{Ablation study.} 
In order to demonstrate the benefits of RidgeSfM's bundle adjustment from \Cref{s:rsfm_bundle}, we compare the performance of the full optimization minimizing $\mathcal{L}^{\text{bundle}}$, and merely minimizing $\sum_{(i,j) \in \mathcal{I}} \mathcal{L}^\text{pose}_{ij}$ that solely aligns the cameras without optimizing the scene 3D points. For skip-rate $k=30$, the full optimization leads to reductions of
49\% / 43\% / 12\% / 44\% in the 
camera rotation error / camera center error / depth RMSE / point cloud RMSE
respectively. 
This clearly demonstrates the benefits of our joint optimization over the latent depth linear codes and camera extrinsics.
\Cref{tbl:ablation} provides a table with full ablation results.

We also consider the effect of replacing the SuperPoint \cite{superpoint} features with classic SIFT \cite{sift} features. The reconstruction quality decreases as expected, although the results are still strong compared to COLMAP. See again \Cref{tbl:ablation}.

\paragraph{Evaluation of the pairwise alignment.}
For completeness, we also consider the task of estimating relative viewpoint change between a pair of images, i.e. the task described in \Cref{s:rsfm_pairwise}. Note that here we compare with small-scale deep methods that cannot operate in the large scale regime of the previous experimental section. 

We closely follow the evaluation protocol of BA-Net \cite{banet}.
We thus consider the set of 2000 pairs of test images from \cite{banet}. Camera movement is generally quite small:  80\% of the ground truth translations are less than 15cm, and 80\% of the ground truth rotations are less than five degrees. We report the rotation error which is the angle between the ground truth relative
camera rotation and the prediction; and the translation error which is the 
distance of the estimated camera center from the ground truth.
Furthermore, RMSE between the estimated and the ground truth depth is reported.

RidgeSfM is compared to other methods in Table \ref{tbl:pairs}. Qualitative results are presented in \Cref{fig:pairs}.
Results indicate that RidgeSfM outperforms other comparable methods in terms of depth accuracy.
The camera errors are slightly higher than BA-Net and DeepV2D. One explanation for this is that those methods are  trained on image pairs with similar statistics to the test set, so they can develop a prior that is biased to predicting small angles of rotation. Another factor to consider is that once an image has been processed once by RidgeSfM, the keypoints and their factor of variation can be stored compactly to be re-used for additional comparisons to other images. The marginal overhead of additional pairwise comparisons is small, compared to methods using dense image comparisons, which is important for scalability.

\section{Conclusions}
We have proposed RidgeSfM, a novel method for estimating structure from motion that marries classic SfM pipelines capable of bundle-adjusting huge image collections with deep reconstructors. 
RidgeSfM's efficient linear parametrization of depth allows to execute both pairwise and scene-wise geometry optimization over a set of sparse matches while simultaneously recovering dense depth in an indirect fashion.
The latter allows our method to collectively align large image collections, which is not possible with the current memory-hungry deep methods.
We perform on par with strongly supervised deep pairwise egomotion estimators and we significantly outpeform a state-of-the-art SfM pipeline on a large-scale bundle adjustment benchmark.

\bibliographystyle{ieee}
\bibliography{arxiv}

\clearpage
\appendix
\section*{Supplementary material}

In what follows we provide additional details about RidgeSfM together with supplementary evaluations. 
Section \ref{s:sup_arch} contains information about the architecture of $B^{\mu/\sigma}$, 
Section \ref{s:sup_pw_rsfm} describes pairwise RidgeSfM in more details including qualitative results,
Section \ref{s:sup_perf} provides performance analysis, and
Section \ref{s:sup_bundle_rsfm_qual} details further qualitative evaluation of RidgeSfM. 

Please see \ \url{https://github.com/facebookresearch/RidgeSfM}\  for videos showing the RidgeSfM reconstructions for a variety of scenes.

\section{U-Net architecture} \label{s:sup_arch}

In Table~\ref{tbl:unet} we give the full architecture of the depth basis predictor $B^{\mu/\sigma}$ from \cref{s:depth_param}. 
The computation cost is 830MFlops (multiply-add compute) operations per image. 
The run time is 70ms per image (14 fps) for a single Intel Core i7 3930K CPU core, or 4ms per image (250fps) with an Nvidia GeForce GTX Titan X. 
The network was trained on the train set of the ScanNet dataset using an SGD optimizer with momentum, with learning rate of $10^{-3}$ decaying ten-fold whenever the losses plateau.

\begin{table*}[ht]
  \caption{{\bf Network architecture} Depth prediction U-Net architecture for input with size 240$\times$320.
  IR$(t)\times n$ denotes a chain of $n$ inverted residual blocks \cite{InvertedResidual} each with expansion factor $t$.  \label{tbl:unet}
  }
  \centering{}%
  \begin{tabular}{lrrc}
  \toprule 
  Layer & Input Features & Output Features & Output Resolution\\
  \midrule
  Conv 4/2      & 3                   & 16              & $120\times160$\\
  Conv 4/2      & 16                  & 32              & $60\times80$\\
  IR$(t=4)\times$2 & 32                  & 32$^{\star}$        & $60\times80$\\
  Conv 4/2      & 32                  & 64              & $30\times40$\\
  IR$(t=4)\times$2 & 64                  & 64$^{\star\star}$    & $30\times40$\\
  Conv 4/2      & 64                  & 96              & $15\times20$\\
  IR$(t=4)\times$2 & 96                  & 96$^{\dagger}$ & $15\times20$\\
  Conv (4,3)/2  & 96                  & 128             & $7\times10$\\
  IR$(t=4)\times$2 & 128                 & 128$^{\ddagger}$ & $7\times10$\\
  Conv (4,3)/2  & 128                 & 160             & $3\times5$\\
  IR$(t=4)\times$4 & 160                 & 160             & $3\times5$\\
  TConv (4,3)/2 & 160                 & 128             & $7\times10$\\
  IR$(t=2)\times$2 & 128$^{\ddagger}$+128 & 128             & $7\times10$\\
  TConv (4,3)/2 & 128                 & 96              & $15\times20$\\
  IR$(t=2)\times$2 & 96$^{\ddagger}$+96  & 96              & $15\times20$\\
  TConv 4/2     & 96                  & 64              & $30\times40$\\
  IR$(t=2)\times$2 & 64$^{\star\star}$+64     & 64              & $30\times40$\\
  TConv 4/2     & 64                  & 32              & $60\times80$\\
  IR$(t=2)\times$2 & 32$^{\star}$+32         & 32              & $60\times80$\\
  Output  & 32                  & 1+32            & $60\times80$\\
  \bottomrule
  \end{tabular}
  \end{table*}

\section{Details on pairwise RidgeSfM} \label{s:sup_pw_rsfm}

Here, we detail the pairwise egomotion estimation by progressive growing of matches, which was briefly outlined in \cref{s:rsfm_pairwise}. In order to increase robustness of the pairwise matching procedure, we employ our algorithm in an iterative fashion that progressively grows a set of inliers taken from a large pool of tentative matches. 

We first extract SuperPoint \cite{superpoint} features from both images. From these we create an initial set of weakly verified matches by looking for keypoints which are $k$-nearest neighbors in the descriptor space and pass the crosschect test. We call this intial set of matches $\mathcal{M}_{ij}^0$. This set will be filtered using a two stage RANSAC process.

We first filter $\mathcal{M}_{ij}^0$ with OpenCV's {\tt findFundamentalMat} function to find the largest subset of matches $\mathcal{M}_{ij}\subset\mathcal{M}_{ij}^0$ consistent with a 3D rotation/translation transform between two images. This initial RANSAC filtering takes place in the two-dimensional image space without using the predicted depth.

We then use the depth predictions $B^{\mu/\sigma}$ to filter $\mathcal{M}_{ij}$ for a second time using the following RANSAC procedure:  
1) Sample an initial subset $\mathcal{M}_{ij}^I$ by picking $M:=3$ matches from $\mathcal{M}_{ij}$ uniformly at random.
2) Solve for the pairwise alignment by minimizing $\mathcal{L}_\text{pw}(\mathcal{M}_{ij}^I)$ using \cref{s:rsfm_pairwise}.
3) Given the current estimate of $\{R_{ij}, T_{ij}, \beta_i, \beta_j\}$, 
we redefine $\mathcal{M}_{ij}^I$ 
to be the set of $M:=\ceil{\alpha M}$ matches\footnote{We set the multplicative growth rate $\alpha=1.2$}
$\{(y_i^m, y_j^m)\} \subset \mathcal{M}_{ij}$
that have the lowest 3D alignment error $\ell_{ij}^m$;
4) Repeat steps (1)-(3) while the maximum alignment error $\ell_{ij}^m$ between 
the matches from $\mathcal{M}_{ij}^I$ is lower than $\epsilon=10cm$.
  
For efficiency, the algorithm can be run in parallel using batching. 
Accumulating the results of a number of runs (e.g. 32), we pick the run such that the inlier 2D keypoints cover the largest number of $10\times 10$ pixel squares in images $I_i$ and $I_j$. We discard any pairs of images where less than 30 such squares are covered as negative matches.

\subsection{Qualitative results of pairwise matching}
\begin{figure*}[ht]
    \newcommand{\figh}{4.8cm}
    \newcommand{\figw}{0.40\linewidth}
    \newcommand{\figcol}[1]{
        \begin{minipage}[c]{\figw}%
        \centering%
        \includegraphics[height=\figh]{pairs_superpoint2/#1}
        \end{minipage}%
    }
    \centering%
    \figcol{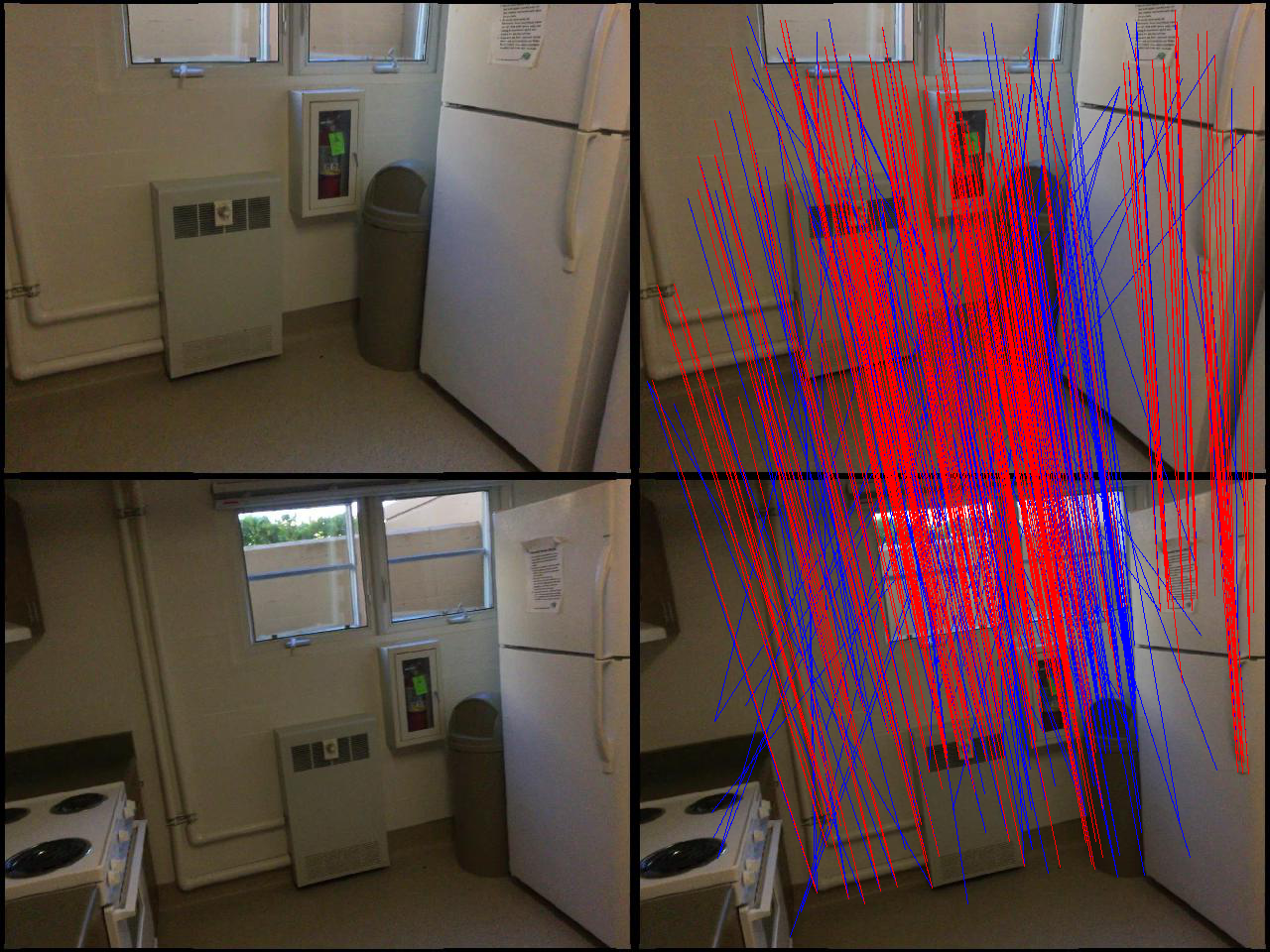}\figcol{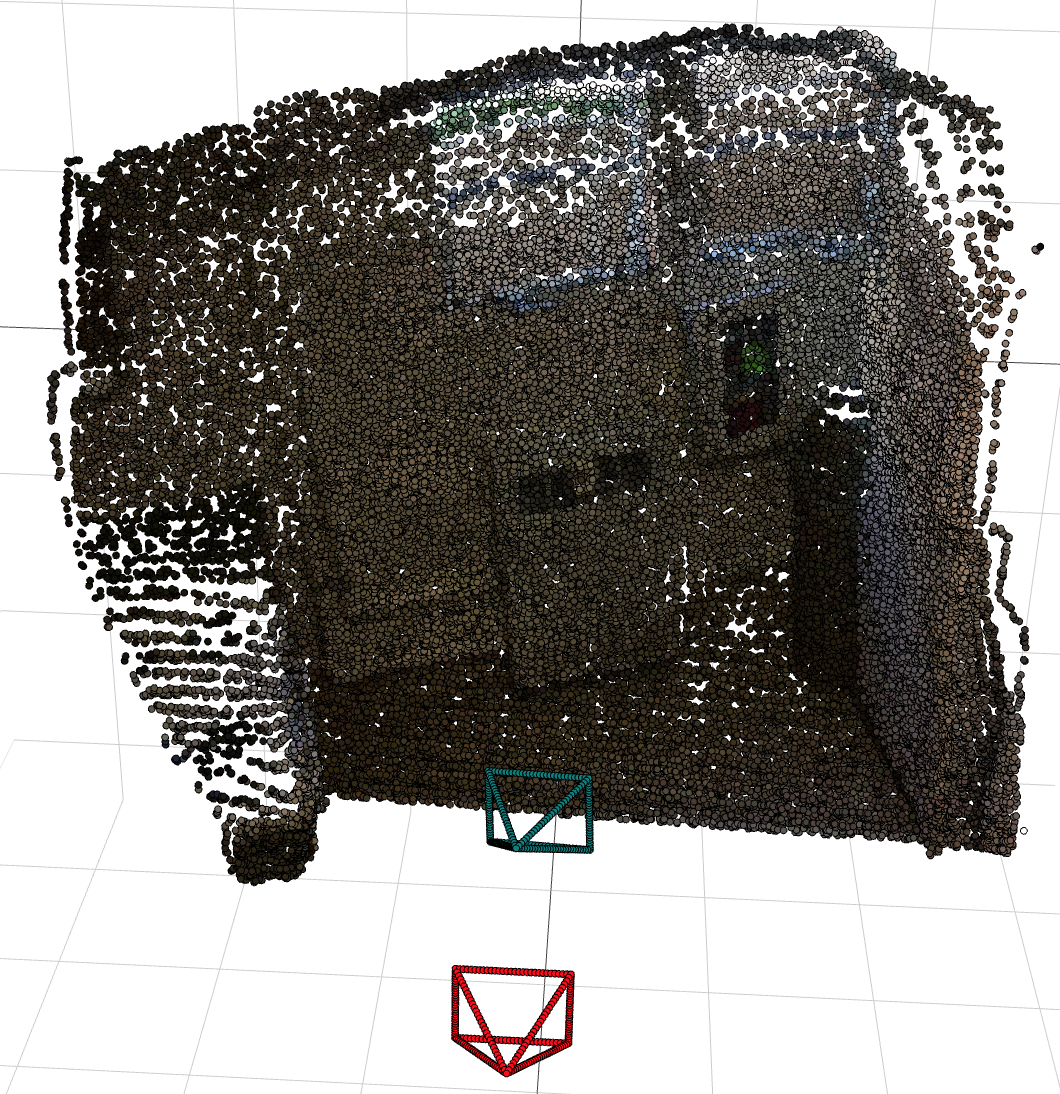}\\
    \figcol{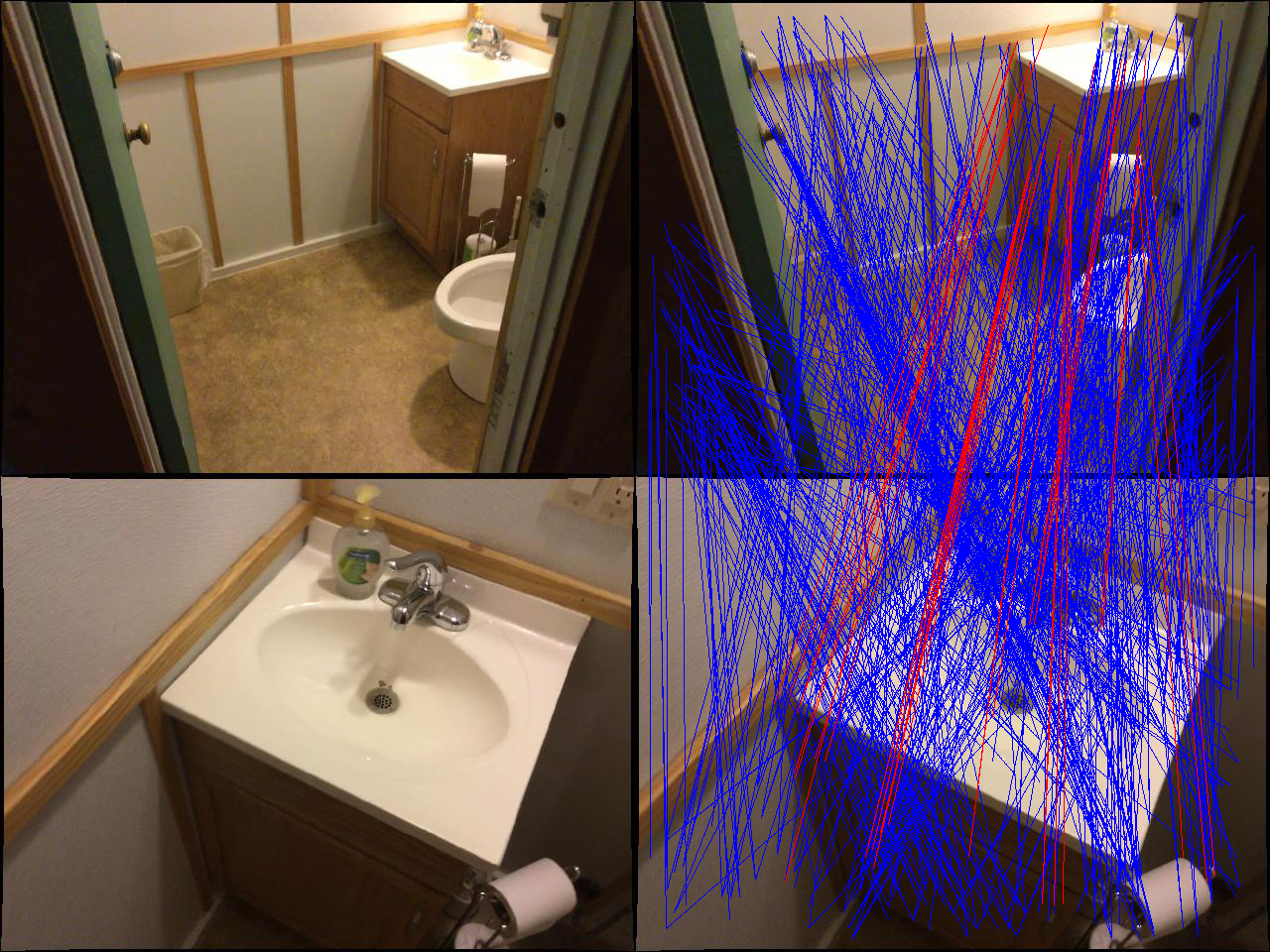}\figcol{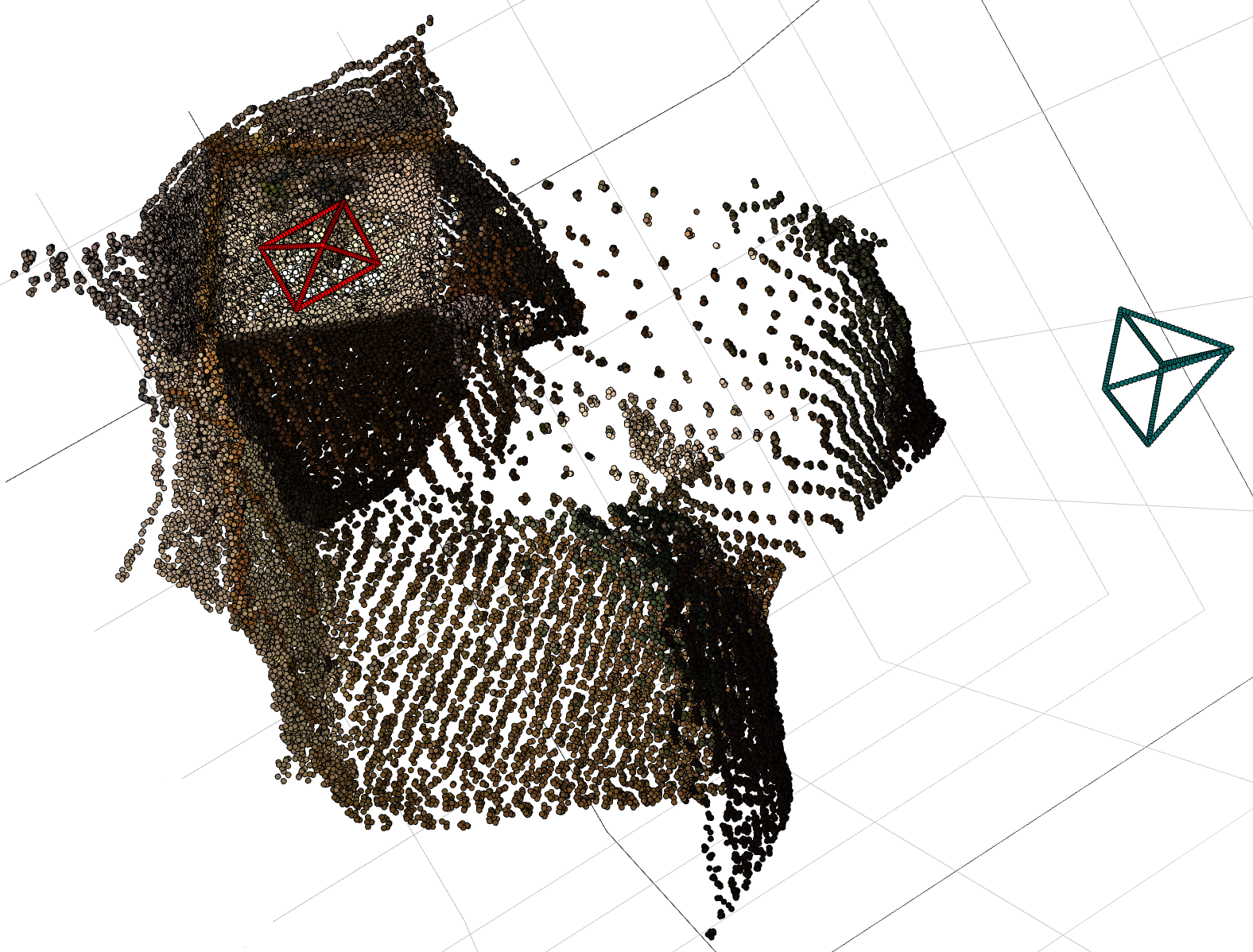}\\
    \figcol{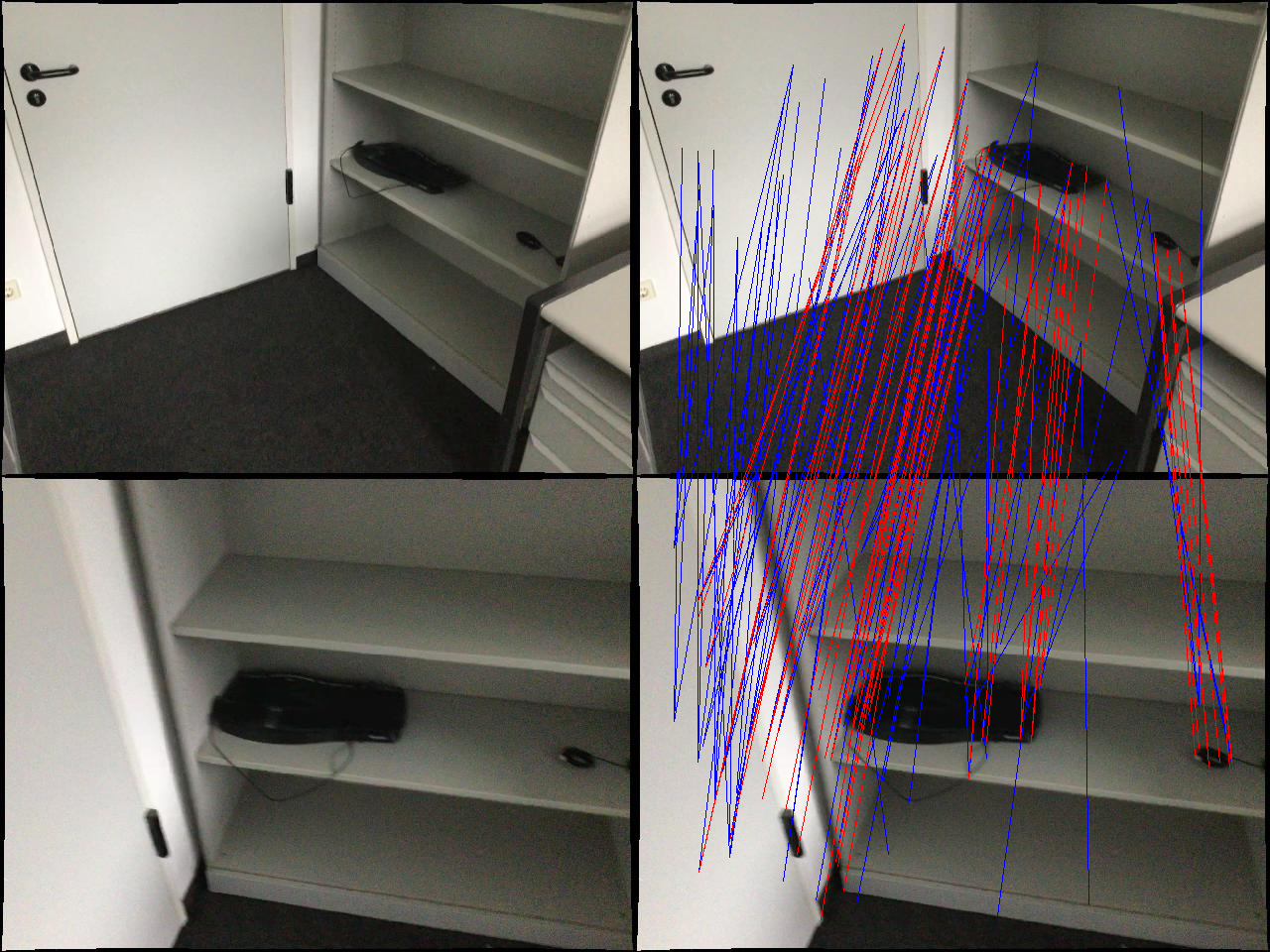}\figcol{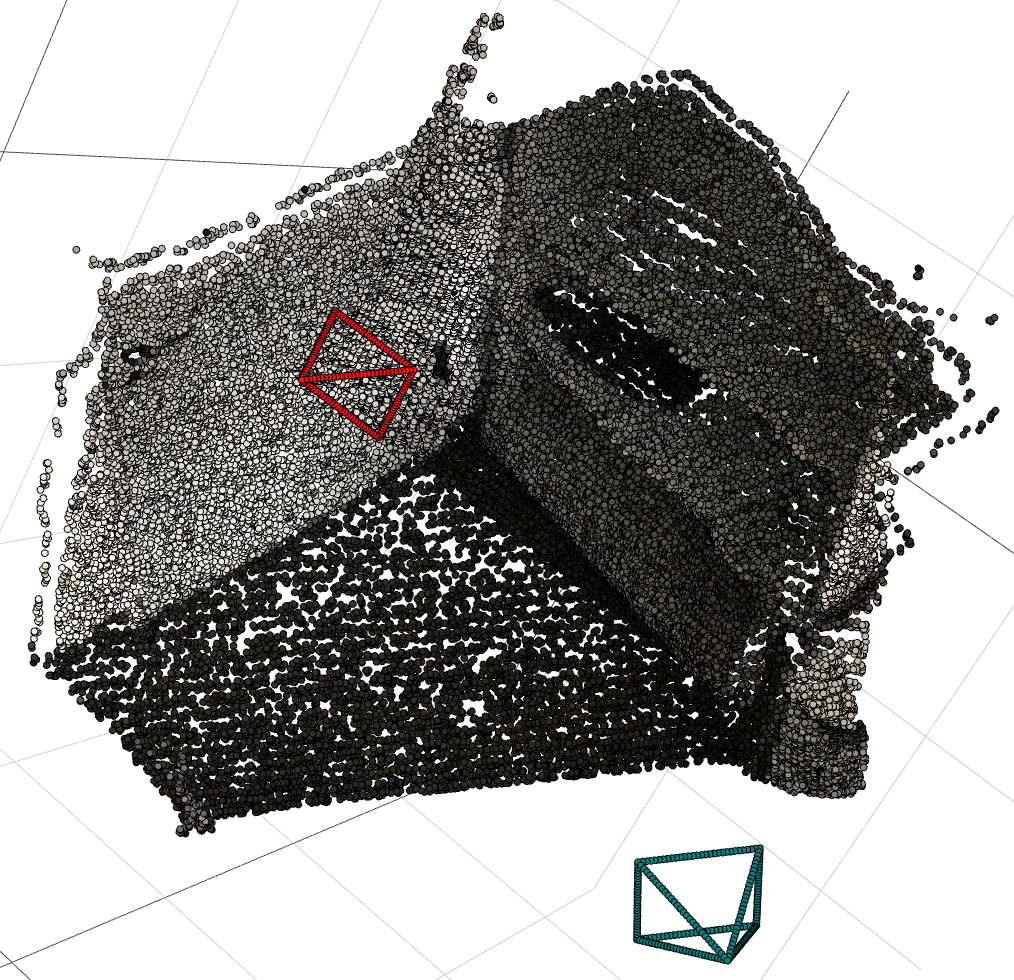}\\
    \figcol{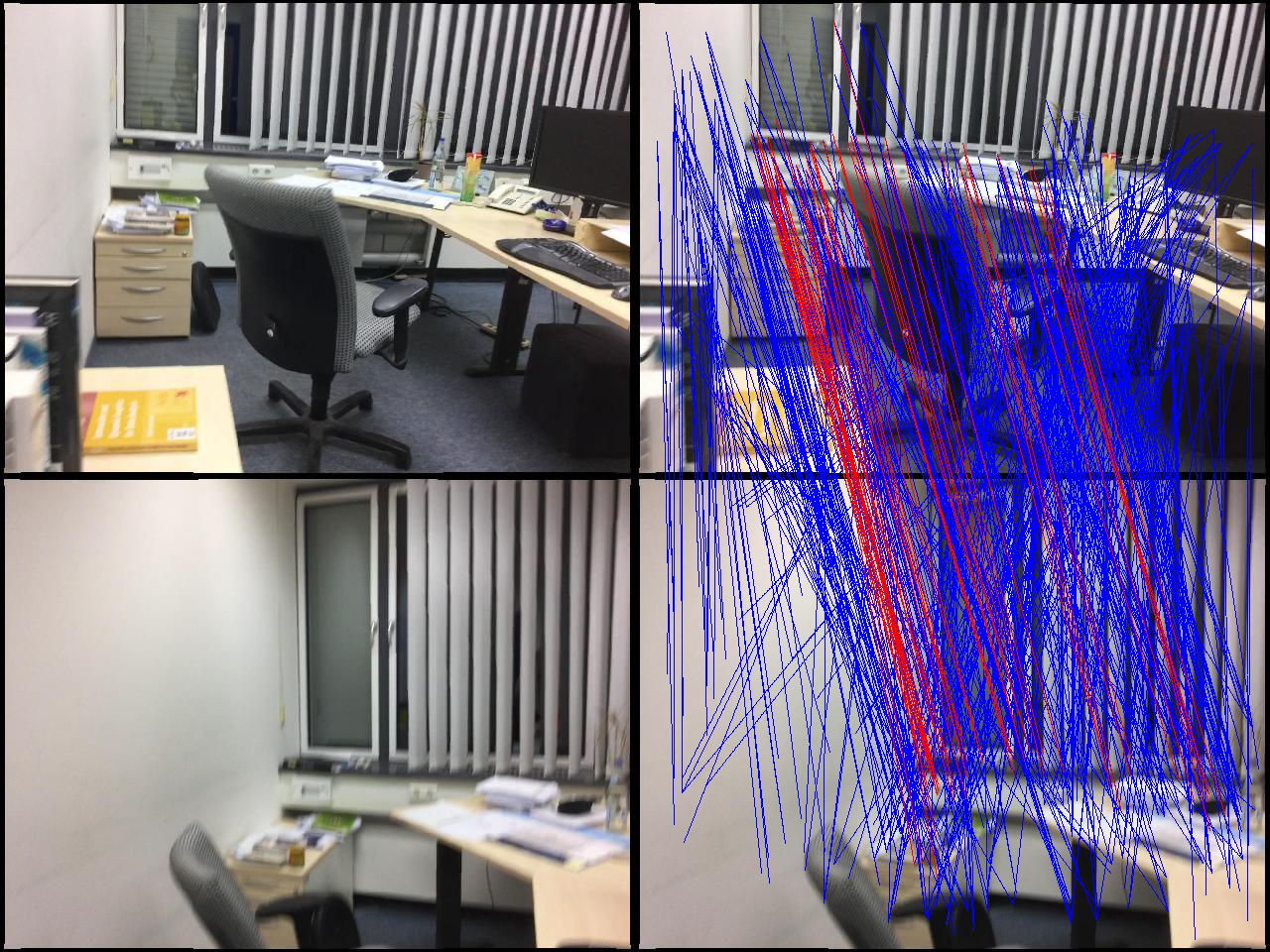}\figcol{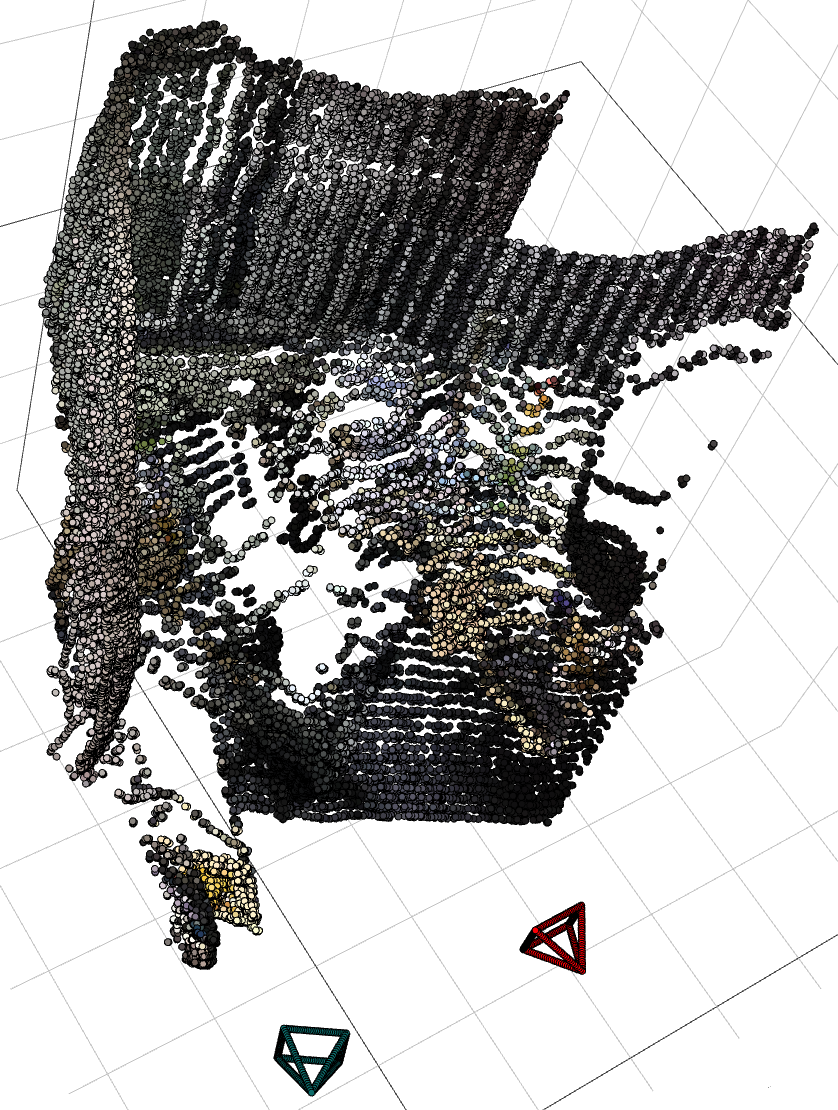}\\
      \caption{ {\bf Pairwise RidgeSfM results} We show here examples of pairwise RidgeSfM matching for images with {\it less} overlap than typically found in the BA-Net \cite{banet} test set. 
         Left: Pairs of images with per-pixel correspondences.  Right: The inferred scene point clouds and cameras - we have plotted 10\% of the pixels.
         The blue lines show initial feature matches. Red matches denote the inliers of the Pairwise RidgeSfM alignment.
        \label{fig:pairs2}}
\end{figure*}

In Figure~\ref{fig:pairs}, we presented qualitative results of the pairwise RidgeSfM on the test set of \cite{banet}. In Figure~\ref{fig:pairs2}, we show pairwise RidgeSfM matches with more substantial camera movements.

\section{Performance analysis}  \label{s:sup_perf}
We ran the reconstruction on a server computer with an Intel Xeon E5-2698 CPU and a Nvidia Quadro GP100 GPU.
Reconstructing a typical ScanNet test scene with 300 frames (frame-skip $k=1$) takes approximately 11 minutes. The bulk of the time is spent doing pairwise RANSAC matching as we try to match 3000 pairs per scene.
In constrast, COLMAP takes approximately 40 minutes.

To see if we could reduce the amount of time spent matching, which also reduces the computational complexity of the bundle adjustment, we considered a lighter weight version of RidgeSfM with only pairwise matches calculated for only 600 pairs. It takes approximately four minutes per scene, whilst maintaining much of RidgeSfM's precision. See Table~\ref{tbl:colmap2} for full details.

\begin{table*}
    \caption{\textbf{RidgeSfM-Light} Quantitative comparison of RidgeSFM, a variant of RidgeSfM with fewer matching operations,  and COLMAP on the ScanNet test set.\label{tbl:colmap2}}
    \centering{}%
    \begin{tabular}{l|cc|cc|cc}
    \toprule
    Method & \multicolumn{2}{c|}{RidgeSfM} &\multicolumn{2}{c|}{RidgeSfM-Light} & \multicolumn{2}{c}{COLMAP}\tabularnewline
    Skip rate & 1 & 3 & 1 & 3 & 1 & 3\tabularnewline
    \midrule
    Average runtime (s) &   650   & 573 &  236 & 220 &  2300& 2083\\
    \midrule   
   Camera rotation (degrees)
    &7.09 &7.84
&8.78 &10.36
    &22.12 &11.17    \\
   Camera center (m)
   &0.296 &0.314
&0.391   &0.439
   &0.973 &0.597\\
   Depth map $L_1$ err. (m)
   &0.221 &0.234
&0.309   &0.350
   &0.941     &0.763     \\
   Depth map RMSE (m)
   &0.305 &0.332
&0.397   &0.455
   &1.138     &1.012      \\
   PCL $L_1$ err. (m)
   &0.209 &0.258
&0.262   &0.356
   &0.647     &0.642     \\
   PCL RMSE (m)
   &0.289 &0.345
&0.343   &0.448
   &0.821      &0.885    \\
   \midrule
   Successful recon. 
   & 100\%       &  100\%       
   & 100\%       &  100\%       
   & 99\%        &  100\%       \\ 
    \bottomrule
    \end{tabular}
    \end{table*}

\section{Videos of RidgeSfM reconstructions} \label{s:sup_bundle_rsfm_qual}

To qualitatively evaluate RidgeSfM reconstructions, we:
\begin{enumerate}
    \item Form a point cloud by projecting every pixel into 3D space using the predicted depth and camera poses.
    \item Each point in the cloud is an element $(x,y,z,r,g,b)\in\mathbb{R}^6$. To make the point cloud more manageable, we sample $10^5$ centroids using K-Means.
    \item We reproject the centroids back into the predicted camera locations using splat rendering.
\end{enumerate} 
Please see \ \url{https://github.com/facebookresearch/RidgeSfM}\  for videos showing the RidgeSfM reconstructions for the first four ScanNet test scenes. 

\subsection{Video from a mobile phone camera} 

To test if RidgeSfM can be applied to scenes outside of the ScanNet dataset, we captured a video of an indoor scene using a mobile phone. The video was resized and cropped to $640\times480$ so that we can apply RidgeSfM exactly as we did for the ScanNet test set. We recycle the the $B^{\mu/\sigma}$ depth prediction network that was trained of the ScanNet training set. 
The mobile phone's camera intrinsics were estimated using OpenCV's \verb calibrateCamera \ function applied to keypoints from a second video of a chess board calibration pattern. The mobile phone has a narrower field of view, with a focal length of 667 compared to 578 for ScanNet.

\subsection{Video from KITTI} 

We also include sample reconstructions for the KITTI dataset \cite{Geiger2012CVPR}.
We trained a depth prediction network on the KITTI depth prediction training set. We then processed videos from the KITTI Visual Odometry dataset. We used the `camera 2' image sequences, cropping the input to RGB images of size $1216\times 320$. For the keypoint detector, we used R2D2 \cite{r2d2} instead of SuperPoint \cite{superpoint}, as R2D2 is trained on photos of outdoor scenes.
    
\begin{figure*}[t] \centering
\includegraphics[width=1\textwidth]{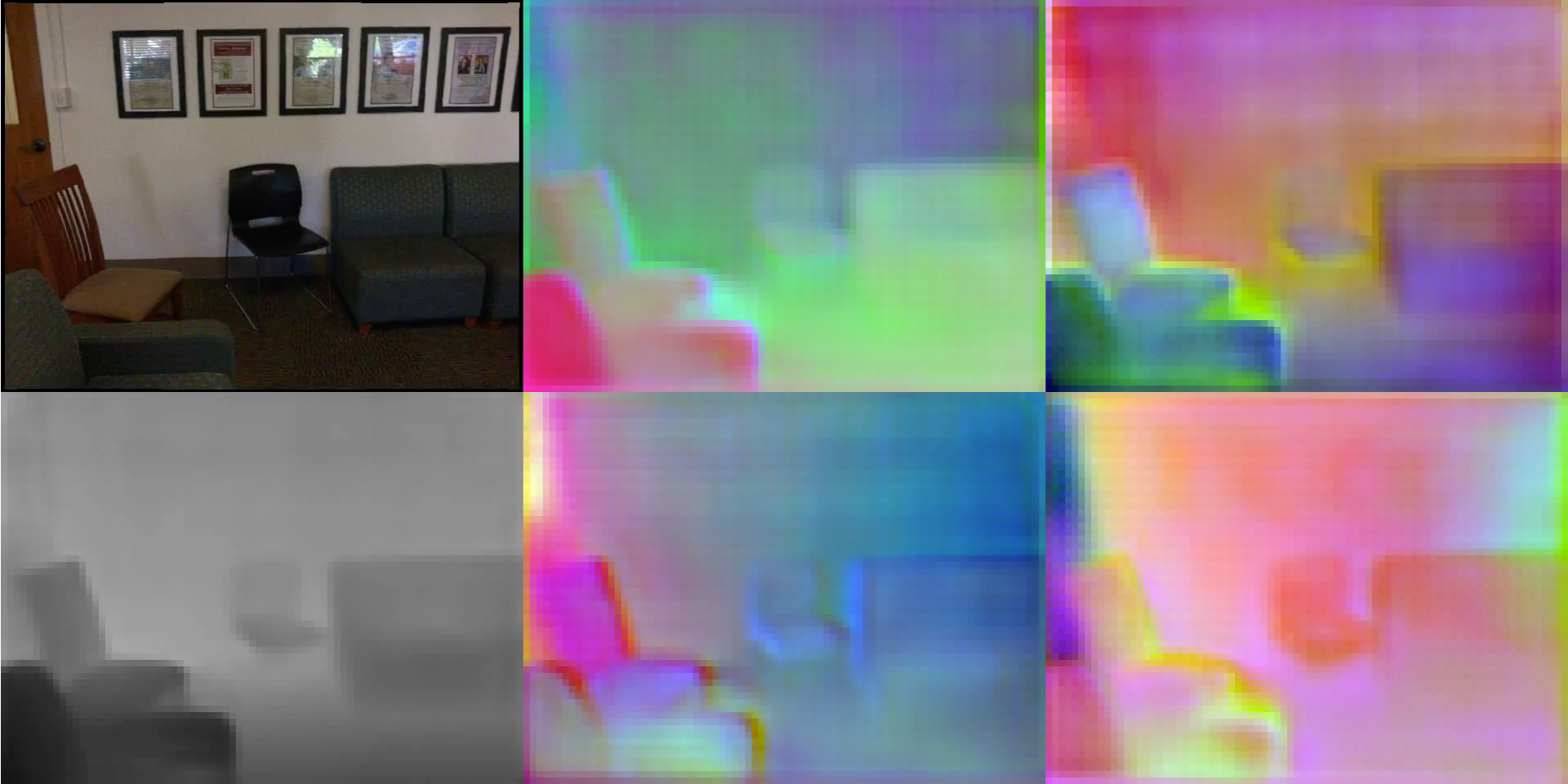}
\caption{{\bf The depth prediction network} Top left: an input image.
Bottom left: the predicted depth.
Middle and right: We use SVD to reduce the 32 factors-of-variation planes down to 12 planes, and display them as 4 RGB images; each of the 4$\times$3 color planes represents one factor of variation.\label{fig:fov12}}
\end{figure*} 

\end{document}